\documentclass[10pt,twocolumn,letterpaper]{article}

\newif\ifarxiv
\arxivtrue  

\ifarxiv
\usepackage[pagenumbers]{cvpr}
\else
\usepackage{cvpr}
\fi

\definecolor{cvprblue}{rgb}{0.21,0.49,0.74}
\usepackage[pagebackref,breaklinks,colorlinks,allcolors=cvprblue]{hyperref}

\usepackage[utf8]{inputenc} 
\usepackage[T1]{fontenc}    
\usepackage{url}            
\usepackage{booktabs}       
\usepackage{amsfonts}       
\usepackage{nicefrac}       
\usepackage{microtype}      
\usepackage{amsmath}
\usepackage{amssymb}
\usepackage{multirow}
\usepackage{colortbl}

\usepackage[accsupp]{axessibility} 

\def\paperID{429} 
\def\confName{CVPR}
\def\confYear{2025}

\title{Point2RBox-v2: Rethinking Point-supervised Oriented Object Detection\\ with Spatial Layout Among Instances}

\author{\textbf{Yi Yu}$^{1*}$,~\textbf{Botao Ren}$^{2}$\thanks{Equal contribution. $^{\dag}$Corresponding author. The work was partly supported by National Natural Science Foundation of China (62306069, 62222607, 72342023), China Postdoctoral Science Foundation (2023M 740602), National Key R\&D Program of China (2022ZD0160100), Jiangsu Funding Program for Excellent Postdoctoral Talent (2023ZB616).}\,\,,~\textbf{Peiyuan Zhang}$^{3*}$,~\textbf{Mingxin Liu}$^{4}$,~\textbf{Junwei Luo}$^{3}$,~\textbf{Shaofeng Zhang}$^{4}$,\\\textbf{Feipeng Da}$^{1}$,~\textbf{Junchi Yan}$^{4}$,~\textbf{Xue Yang}$^{4\dag}$\\
$^{1}$Southeast University \hspace{2pt} $^{2}$Tsinghua University \hspace{2pt}
$^{3}$Wuhan University \hspace{2pt} $^{4}$Shanghai Jiao Tong University\\
{\vspace{-2pt}\tt\small \{yuyi,dafp\}@seu.edu.cn rbt22@mails.tsinghua.edu.cn \{peiyuanzhangwhu,luojunwei\}@whu.edu.cn}\\
{\vspace{-2pt}\tt\small \{liumingxin,zhangshaofeng,yanjunchi,yangxue-2019-sjtu\}@sjtu.edu.cn}\\
{\tt\small \url{https://github.com/VisionXLab/point2rbox-v2}}}

\begin{document}
\maketitle
\begin{abstract}
With the rapidly increasing demand for oriented object detection (OOD), recent research involving weakly-supervised detectors for learning OOD from point annotations has gained great attention. 
In this paper, we rethink this challenging task setting with the layout among instances and present Point2RBox-v2. 
At the core are three principles: 
\textbf{1) Gaussian overlap loss.} It learns an upper bound for each instance by treating objects as 2D Gaussian distributions and minimizing their overlap. 
\textbf{2) Voronoi watershed loss.} It learns a lower bound for each instance through watershed on Voronoi tessellation. 
\textbf{3) Consistency loss.} It learns the size/rotation variation between two output sets with respect to an input image and its augmented view. 
Supplemented by a few devised techniques, e.g. edge loss and copy-paste, the detector is further enhanced.
To our best knowledge, Point2RBox-v2 is the first approach to explore the spatial layout among instances for learning point-supervised OOD. Our solution is elegant and lightweight, yet it is expected to give a competitive performance especially in densely packed scenes: 62.61\%/86.15\%/34.71\% on DOTA/HRSC/FAIR1M.
\end{abstract}    
\vspace{-6pt}
\section{Introduction}
\label{sec:intro}

Emerged as an essential task in computer vision, oriented object detection (OOD) has become a prominent demand in autonomous driving \cite{feng2021deep}, aerial images \cite{xia2018dota, liu2017hrsc, yang2018automatic, yang2022arbitrary, fu2020rotation}, scene text \cite{liao2018rotation, liu2018fots, zhou2017east}, retail scenes \cite{goldman2019precise, pan2020dynamic}, industrial inspection \cite{liu2020data, wu2022pcbnet}, and more.

\begin{figure}[t!]
\setlength{\abovecaptionskip}{1.2mm}
\centering
\includegraphics[width=\linewidth]{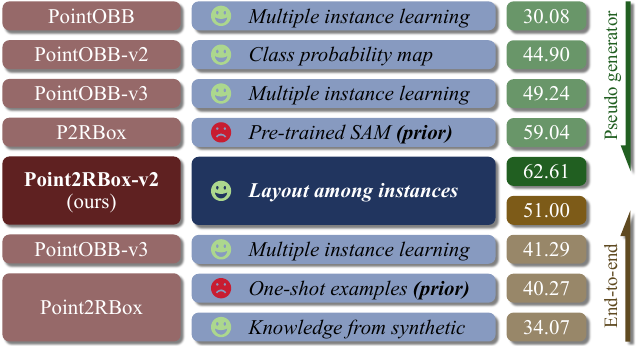}
\caption{Related methods, their principles for knowledge mining, whether using additional priors, and performance on DOTA-v1.0.}
\label{fig:intro}
\vspace{-6pt}
\end{figure}

Manual annotations are essential to teach the detector new concepts of visual objects. Early research is often supervised by rotated bounding boxes (RBoxes), where annotations align with the desired output. While this approach has demonstrated promising results, RBox labels are considerably expensive. The cost of annotating each RBox is approximately 36.5\% higher than a horizontal bounding box (HBox) and 104.8\% higher than a point \cite{yu2024point2rbox}. This highlights the importance of utilizing coarser labels for weakly-supervised oriented detection. Recent advancements in HBox-supervised OOD, particularly with H2RBox \cite{yang2023h2rbox} and H2RBox-v2 \cite{yu2023h2rboxv2}, have shown promise in bridging the gap between HBox- and RBox-supervised methods, thereby lessening the dependence on labor-intensive RBox labeling. 

\begin{figure*}[t!]
\setlength{\abovecaptionskip}{1.2mm}
\centering
\includegraphics[width=0.96\linewidth]{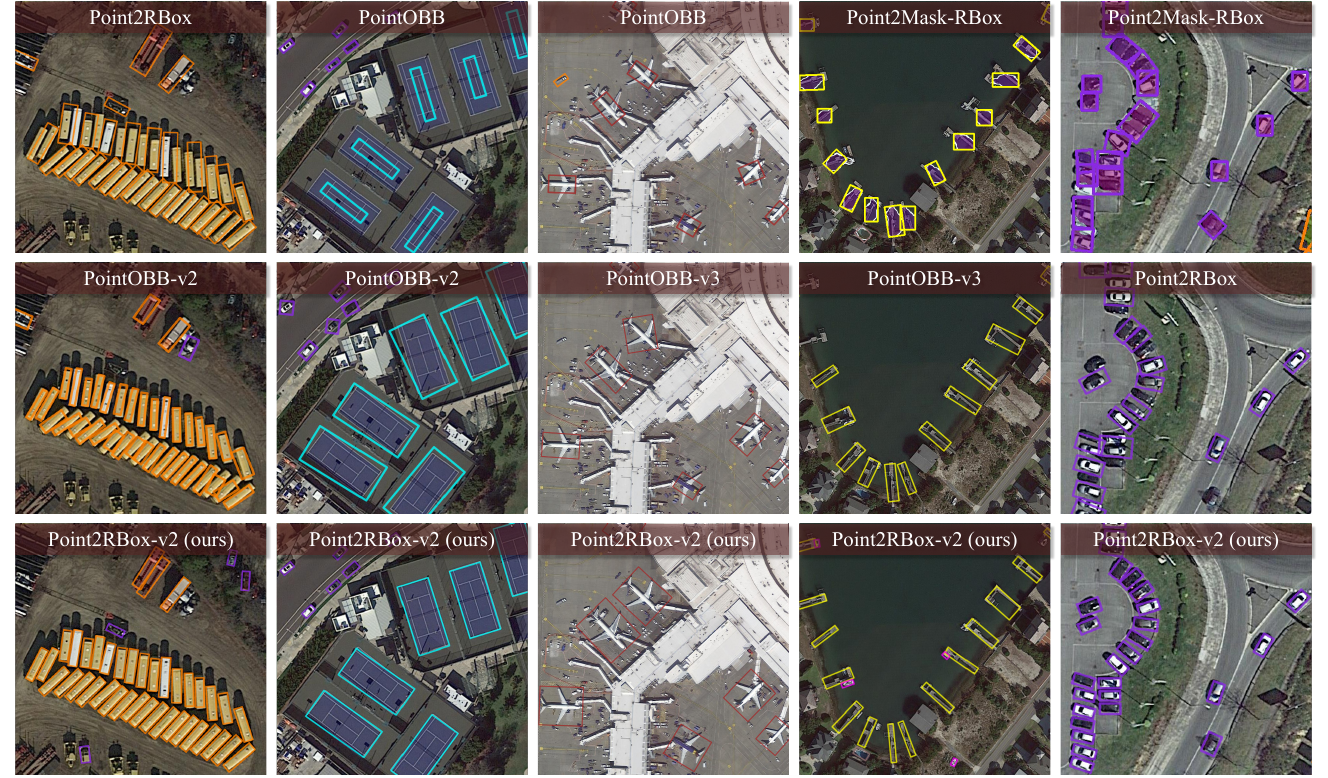}
\caption{Visual comparisons with state-of-the-art methods including: Point2Mask (2023) \cite{li2023point2mask}, PointOBB (2024) \cite{luo2024pointobb}, PointOBB-v2 (2025) \cite{ren2024pointobbv2}, PointOBB-v3 (2025) \cite{zhang2025pointobbv3}, and Point2RBox (2024) \cite{yu2024point2rbox}. The boxes detected by our method (last row) wrap the objects more tightly.}
\label{fig:vis}
\vspace{-6pt}
\end{figure*}

Following HBox-supervised methods, the more challenging point-supervised OOD opens up a new avenue. Several innovative approaches have emerged in the past year (see Fig. \ref{fig:intro}):
\textbf{1) Point-prompt OOD.} P2RBox \citep{cao2023p2rbox}, PMHO \citep{zhang2024pmho}, and PointSAM \citep{liu2024pointsam} employ the zero-shot ability of SAM \citep{kirillov2023segment}.
\textbf{2) Pseudo generation.} PointOBB \cite{luo2024pointobb} and its v2/v3 version \cite{ren2024pointobbv2, zhang2025pointobbv3} use multiple instance learning and class probability map for RBox generation. 
\textbf{3) Knowledge combination.} Point2RBox \cite{yu2024point2rbox} learns from one-shot examples.
However, existing methods overlook the spatial relationships between objects. This brings us to the motivation of this paper.

\textbf{Motivation.} While point-supervised OOD has gained attention, the utilization of relationships among instances remains absent in the literature. For example, the arrangement of vehicles in a parking lot can serve as effective constraints for learning their size and orientation. This approach may yield valuable insights into point-supervised OOD, particularly in densely packed scenes where current methods face significant challenges. Can we leverage the spatial layout of objects to enhance point-supervised OOD? In this paper, we dive into this idea and seek to answer this question.

\textbf{Highlights.} \textbf{1)} Point2RBox-v2 is proposed for point-supervised OOD, advancing the state of the art as displayed in Fig.~\ref{fig:vis} and Tables \ref{tab:exp_dota}-\ref{tab:exp_other}. 
\textbf{2)} We propose novel and elegant losses to enforce constraints from the spatial layout among instances based on Gaussian overlap and Voronoi tessellation \cite{aurenhammer1991voronoi}. 
\textbf{3)} Other modules are devised (i.e. edge loss, consistency loss, copy-paste) to further enhance the method.

\textbf{Contributions.} \textbf{1)} To our best knowledge, this work is the first attempt to learn point-supervised OOD from the layout among objects, where we propose novel principles based on Gaussian overlap and Voronoi tessellation.
\textbf{2)} The training pipeline and detailed implementation are elucidated, with necessary modules (i.e. edge loss, consistency loss, copy-paste) incorporated. The source code will be made publicly available.
\textbf{3)} Extensive experiments demonstrate that leveraging the spatial layout of instances can significantly advance the state of the art, surpassing other alternatives in accuracy.

\section{Related Work}
\label{sec:related}

\begin{figure*}[t]
\setlength{\abovecaptionskip}{1.2mm}
\centering
\includegraphics[width=0.98\linewidth]{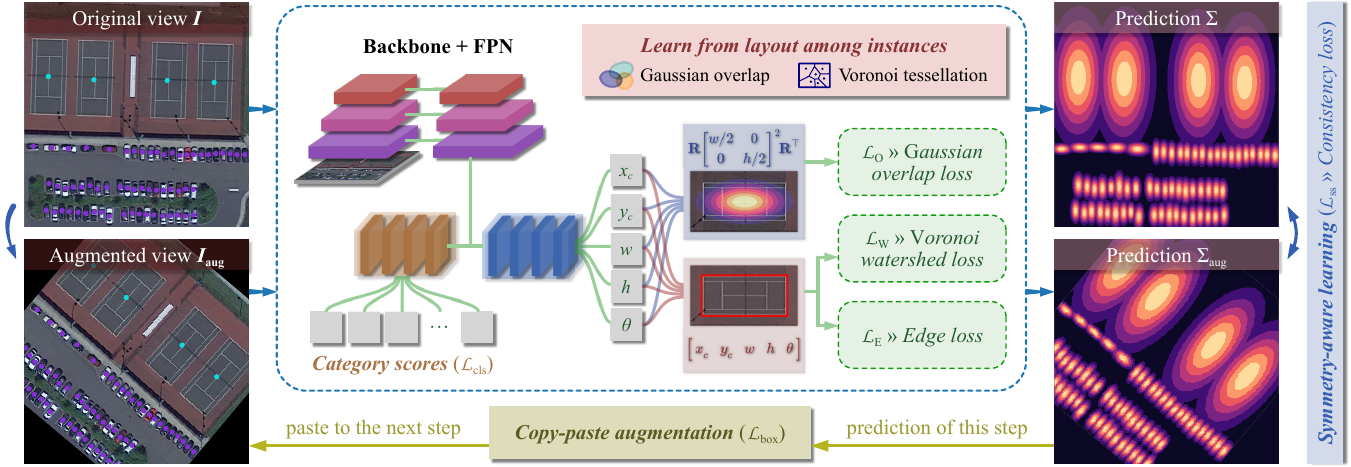}
\caption{The training pipeline of Point2RBox-v2. Gaussian overlap loss and Voronoi watershed loss utilize the spatial layout (see Fig. \ref{fig:loss}), while edge loss (see Sec. \ref{sec:method-le}), symmetry-aware learning (see Sec. \ref{sec:method-lss}), and copy-paste (see Sec. \ref{sec:method-cp}) further enhance the method.}
\label{fig:arch}
\vspace{-6pt}
\end{figure*}

\subsection{RBox-supervised Oriented Detection} 

In addition to horizontal detection \citep{zhao2019object,liu2020deep}, oriented object detection (OOD) \citep{yang2018automatic,wen2023comprehensive} has received extensive attention. Representative works include anchor-based detector Rotated RetinaNet \citep{lin2017focal}, anchor-free detector Rotated FCOS \citep{tian2019fcos}, and two-stage solutions, e.g. RoI Transformer \citep{ding2018learning}, Oriented R-CNN \citep{xie2021oriented}, and ReDet \citep{han2021redet}. Some research enhances the detector by exploiting alignment features, e.g. R$^3$Det \citep{yang2021r3det} and S$^2$A-Net \citep{han2022align}. The angle regression may face boundary discontinuity and remedies are developed, including modulated losses \citep{yang2019scrdet, yang2022scrdet++, qian2021rsdet} that alleviate loss jumps, angle coders \citep{yang2020arbitrary, yang2021dense, yang2022arbitrary, yu2023psc} that convert the angle into boundary-free coded data, and Gaussian-based losses \citep{yang2021rethinking, yang2021learning, yang2023detecting, yang2023kfiou, murrugarra2024probabilistic} transforming rotated bounding boxes into Gaussian distributions. RepPoint-based methods \citep{yang2019reppoints, hou2022grep, li2022oriented} provide alternatives that predict a set of points that bounds the spatial extent of an object. LMMRotate \cite{li2025simple} is a new paradigm of OOD based on multimodal language model and performs object localization through autoregressive prediction.

\subsection{Point-supervised Oriented Detection}

Recently, several methods for point-supervised oriented detection have been proposed: \textbf{1)} P2RBox \citep{cao2023p2rbox}, PMHO \citep{zhang2024pmho}, and PointSAM \citep{liu2024pointsam} propose oriented detection with point prompts by employing the zero-shot Point-to-Mask ability of SAM \citep{kirillov2023segment}. \textbf{2)} Point2RBox \citep{yu2024point2rbox} introduces an end-to-end approach based on knowledge combination. \textbf{3)} PointOBB \citep{luo2024pointobb, zhang2025pointobbv3} achieves RBox generation through scale consistency and multiple instance learning. \textbf{4)} PointOBB-v2 \cite{ren2024pointobbv2} learns a class probability map to generate pseudo RBox labels.

Among these methods, P2RBox, PMHO, and PointSAM rely on the SAM model pre-trained on massive labeled datasets, whereas Point2RBox requires one-shot examples for each category. In contrast, PointOBB series do not use many priors, but they necessitate two-stage training. 

\subsection{Other Weakly-supervised Settings} 

Compared to Point-to-RBox, some other settings have been better studied. These methods are potentially applicable to our Point-to-RBox task setting by using a cascade pipeline, such as Point-to-HBox-to-RBox. In our experiment, cascade pipelines powered by state-of-the-art approaches are also compared. Here, representative works are introduced.

\textbf{HBox-to-RBox.} H2RBox \citep{yang2023h2rbox} establishes a paradigm that limits the object to a few candidate angles through geometric constraint from HBoxes, with a self-supervised branch eliminating the undesired results. An enhanced version H2RBox-v2 \citep{yu2023h2rboxv2} is proposed to leverage the reflection symmetry of objects to further boost the accuracy. EIE-Det~\citep{wang2024explicit} uses an explicit equivariance branch for learning rotation consistency, and an implicit equivariance branch for learning position, aspect ratio, and scale consistency. KCR \cite{zhu2023knowledge} combines RBox- and HBox-annotated datasets for transfer learning. Some studies~\citep{iqbal2021leveraging,sun2021oriented} use additional annotated data for training, which are also attractive but less general.

\textbf{Point-to-HBox.} Several related approaches have been developed, including: \textbf{1)} P2BNet \citep{chen2022pointtobox} samples box proposals of different sizes around the labeled point and classify them to achieve point-supervised horizontal object detection. \textbf{2)} PSOD \citep{gao2022weakly} achieves point-supervised salient object detection using an edge detector and adaptive masked flood fill.

\textbf{Point-to-Mask.} Point2Mask \citep{li2023point2mask} is proposed to achieve panoptic segmentation using single point annotation per target. SAM (Segment Anything Model) \citep{kirillov2023segment} produces object masks from input point/HBox prompts. Though RBoxes can be obtained from the segmentation mask by finding the minimum circumscribed rectangle, such a complex pipeline can be less cost-efficient and perform worse \citep{yang2023h2rbox, yu2023h2rboxv2}.

\section{Method}
\label{sec:method}

\subsection{Overview and Preliminary}
\label{sec:method-arch}

\begin{figure*}[t]
\setlength{\abovecaptionskip}{1.2mm}
\centering
\includegraphics[width=0.96\linewidth]{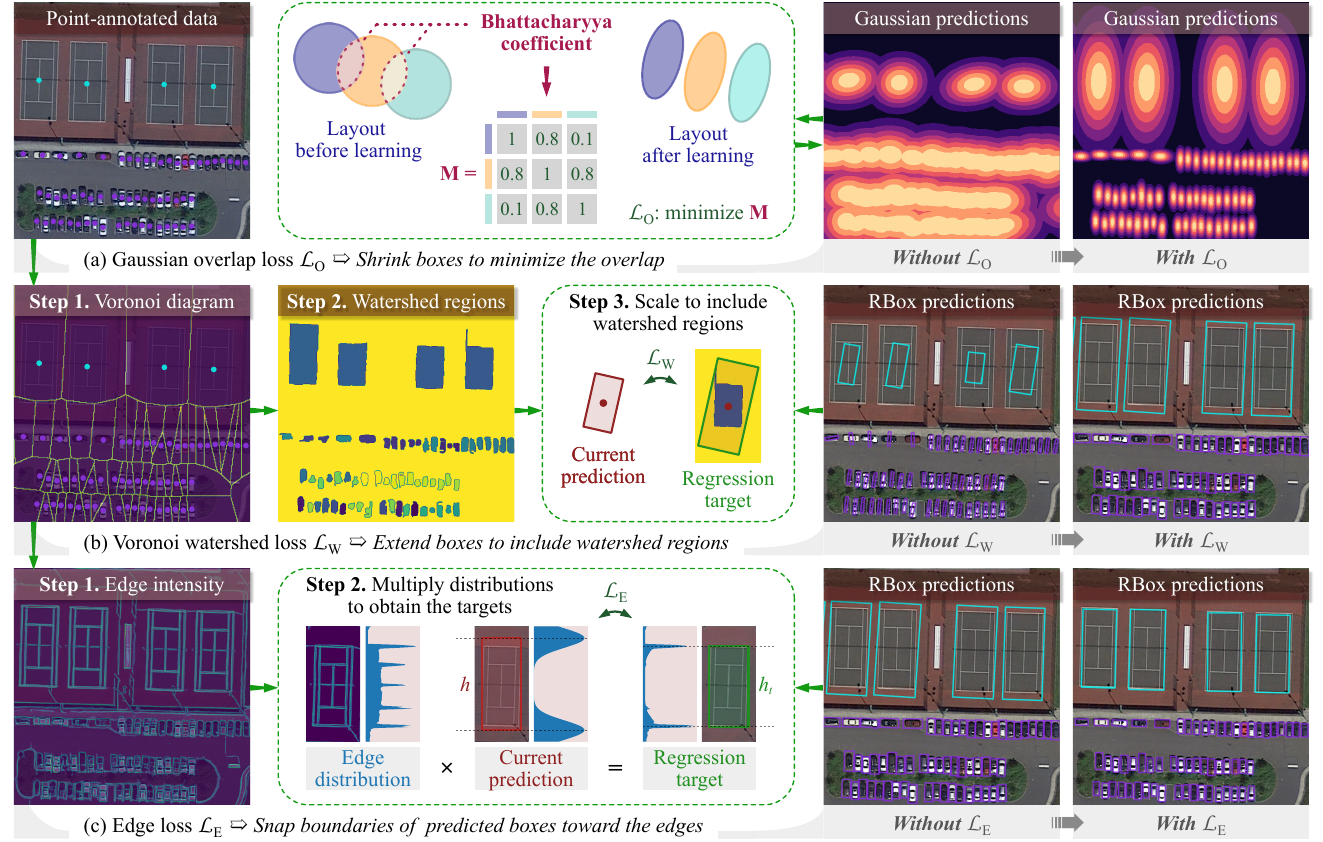}
\caption{To illustrate the procedure of the three newly proposed loss functions and their impact on the learning results. (a) Gaussian overlap loss (see Sec. \ref{sec:method-lo}). (b) Voronoi watershed loss (see Sec. \ref{sec:method-lw}). (c) Edge loss (see Sec. \ref{sec:method-le}).}
\label{fig:loss}
\vspace{-6pt}
\end{figure*}

An overview of Point2RBox-v2 is illustrated in Fig.~\ref{fig:arch}. The network is based on ResNet50 \cite{he2016deep} backbone, FPN \cite{Lin2017Feature} head, and PSC \cite{yu2024boundary} angle coder. Objects of varying sizes are typically assigned to different FPN layers based on the scale. However, points lack size, so we assign them all to the P3 layer (stride = 8). Assume the detector $f_\text{nn}\left ( \cdot \right ) $ maps an image $I$ to a set of RBoxes as detection results:
\begin{equation} \setlength\abovedisplayskip{6pt} \setlength\belowdisplayskip{6pt}
\left(
 x_c, y_c, w, h, \theta
\right) = f_\text{nn}\left ( I \right )
\end{equation}

Equivalently, oriented objects can also be represented by 2D Gaussian distributions $\mathcal{N}(\mu, \Sigma)$ \cite{yang2023detecting}:
\begin{equation} \setlength\abovedisplayskip{6pt} \setlength\belowdisplayskip{6pt}
\begin{Bmatrix}
\mu=\begin{bmatrix}
x_c & y_c
\end{bmatrix},
&
\Sigma=\mathbf{R}\begin{bmatrix}
 w/2 & 0 \\
 0 & h/2
\end{bmatrix} ^{2} \mathbf{R}^\top 
\end{Bmatrix}
\end{equation}
where 
\begin{equation} \setlength\abovedisplayskip{6pt} \setlength\belowdisplayskip{6pt}
\mathbf{R}=\begin{bmatrix}
 \cos \theta & -\sin \theta \\
 \sin \theta & \cos \theta
\end{bmatrix}
\label{equ:rotation}
\end{equation}

At the core of Point2RBox-v2 is to utilize the constraints from the layout, which is achieved by Gaussian overlap loss (Sec. \ref{sec:method-lo}) and Voronoi watershed loss (Sec. \ref{sec:method-lw}). These two losses effectively limit the size and rotation of objects to a reasonable range. Upon that, edge loss (Sec. \ref{sec:method-le}) aligns the bounding box with the edge of objects to improve the accuracy. Incorporated with symmetry-aware learning (Sec. \ref{sec:method-lss}) and copy-paste augmentation (Sec. \ref{sec:method-cp}), we achieve a stable and high-accuracy solution for point-supervised OOD. In subsequent subsections, these modules are detailed. 

\subsection{Gaussian Overlap Loss}
\label{sec:method-lo}

As mentioned, oriented objects can be represented by 2D Gaussian distributions $\mathcal{N}(\mu, \Sigma)$. The overlap volume between two distributions $\mathcal{N}_1(\mu_1, \Sigma_1)$ and $\mathcal{N}_2(\mu_2, \Sigma_2)$ can be approximated by the Bhattacharyya coefficient \cite{yang2023detecting} as:
\begin{equation} \setlength\abovedisplayskip{6pt} \setlength\belowdisplayskip{6pt}
B\left(\mathcal{N}_1, \mathcal{N}_2 \right) = \exp\!\left( - \frac{1}{8} \mu^\top \Sigma^{-1} \mu \right) \!\cdot\! \frac{|\Sigma_1|^{1/4} |\Sigma_2|^{1/4}}{| \Sigma |^{1/2}}
\end{equation}
where $\mu = \mu_2 - \mu_1$, $\Sigma = \frac{1}{2}(\Sigma_1 + \Sigma_2)$, $ |\Sigma| $ denotes the determinant of the covariance matrix.

Based on the above equation, we build a Gaussian overlap matrix $\mathbf{M} \in \mathbb{R}^{N\times N}$ for each scene image as:
\begin{equation} \setlength\abovedisplayskip{6pt} \setlength\belowdisplayskip{6pt}
\mathbf{M}_{i,j} = B\left ( \mathcal{N}_i, \mathcal{N}_j \right )
\end{equation}
where $i,j = 1, 2, \dots, N$; $N$ is the distribution count (or the instance count) within one training image.

The Gaussian overlap loss can then be expressed as:
\begin{equation} \setlength\abovedisplayskip{6pt} \setlength\belowdisplayskip{6pt}
\mathcal{L}_O = \frac{1}{N} \sum_{i \neq j}^{}  \left ( \mathbf{M}_{i,j} \right )
\end{equation}
where $i \neq j$ omits diagonal elements.
With this loss, the detector learns to arrange instances (see Fig. \ref{fig:loss}a) based on the mutual exclusivity among instances.

\subsection{Voronoi Watershed Loss}
\label{sec:method-lw}

A Voronoi diagram \cite{aurenhammer1991voronoi} is a partitioning of a space based on a set of points. Point annotations can be effectively utilized to calculate a Voronoi diagram, where a distinct polygon region is assigned to each point-annotated instance. The watershed algorithm \cite{vincent1991watersheds}, on the other hand, is a region segmentation technique that treats the intensity of pixels as a topographic surface, identifying regions as ``catchment basins''.

The calculation of Voronoi ridges can be formulated as:
\begin{equation} \setlength\abovedisplayskip{6pt} \setlength\belowdisplayskip{6pt}
V = \mathit{Voronoi} \left (X \right ) 
\end{equation}
where $X$ are the annotated points within a training image; $V$ are the output Voronoi ridges (pixel coordinates).

Interestingly, we find that Voronoi diagrams can be utilized as initial markers for watershed to obtain a region for each instance (see Fig. \ref{fig:loss}b). In concrete terms, the points $X$ can be employed as foreground markers, while the Voronoi ridges $V$ can act as background boundaries:
\begin{equation} \setlength\abovedisplayskip{6pt} \setlength\belowdisplayskip{6pt}
S = \mathit{Watershed} \left (I, X, V \right ) 
\end{equation}
where $S$ are the output basin regions (pixel coordinates) corresponding to each annotated instance.
By rotating $S$ to align with the direction of the current prediction, the regression target of width and height can be expressed as:
\begin{equation} \setlength\abovedisplayskip{6pt} \setlength\belowdisplayskip{6pt}
\begin{bmatrix}
w_t \\
h_t
\end{bmatrix} = 2\max \left |  \mathbf{R}^\top
\left ( S- \begin{bmatrix}
x_c \\
y_c
\end{bmatrix} \right )  \right | 
\end{equation}
where $\left(
 x_c, y_c, w, h, \theta
\right)$ is the current prediction; $\mathbf{R}$ is defined by Eq. (\ref{equ:rotation}); $w_t$ and $h_t$ are detached to stop the gradient.

Afterward, the Voronoi watershed loss to regress the width and height of objects can be calculated as:
\begin{equation} \setlength\abovedisplayskip{6pt} \setlength\belowdisplayskip{6pt}
\mathcal{L}_W = L_\text{GWD}\!\left (\! \begin{bmatrix}
 w/2 & 0 \\
 0 & h/2
\end{bmatrix} ^{2}\!,\begin{bmatrix}
 w_t/2 & 0 \\
 0 & h_t/2
\end{bmatrix} ^{2} \right ) 
\end{equation}
where $L_\text{GWD}\left (\cdot\right )$ is Gaussian Wasserstein Distance Loss \cite{yang2023detecting}.

\subsection{Edge Loss}
\label{sec:method-le}

The above two losses have limited the size to a reasonable range. To make it more accurate, we propose the edge loss to snap the boundaries toward the edges (see Fig. \ref{fig:loss}c).

First, a region $P \in \mathbb{R}^{(2K+1)\times (2K+1)}$ around each predicted RBox is extracted via Rotated RoI Align \cite{he2017maskrcnn}:
\begin{equation} \setlength\abovedisplayskip{6pt} \setlength\belowdisplayskip{6pt}
P = \mathit{RoIAlign}\left ( \mathit{E}\left ( I \right ) ,\left ( x_c, y_c, \beta w, \beta h, \theta \right )  \right ) 
\end{equation}
where $\mathit{E}\left ( \cdot \right )$ is the edge detection function \cite{soria2023teed}. We set $K=24$ and $\beta=1.6$ in our experiments (see Table \ref{tab:abl_edgeparam}).

By calculating the sum of each row of $P$, the edge distribution in $y$ direction can be obtained as:
\begin{equation} \setlength\abovedisplayskip{6pt} \setlength\belowdisplayskip{6pt}
\mu_i = \sum_{j=1}^{2K+1} (P_{(K+1-i),j} + P_{(K+1+i),j})
\end{equation}
where $i = 1, 2, \dots, K$, indicating that the upper half of $P$ is reversed and added to the lower half.

Meanwhile, the current prediction of the edge can also be softened into a distribution as: 
\begin{equation} \setlength\abovedisplayskip{6pt} \setlength\belowdisplayskip{6pt}
\lambda_i = \exp\left(-\frac{(i - K / \beta)^2}{2{\sigma_E}^2}\right)
\end{equation}
where $i = 1, 2, \dots, K$; $\sigma_E$ is set to 6. Note that we crop $P$ based on the predicted RBox, thus $\lambda$ is always the same.

Multiplying the two distributions yields the target:
\begin{equation} \setlength\abovedisplayskip{6pt} \setlength\belowdisplayskip{6pt}
h_t = \frac{\beta h}{K} \arg \max (\mu \times \lambda)
\end{equation}
where $h_t$ is the regression target of height. Likewise, the width target $w_t$ is calculated along $x$ direction.

The edge loss is then obtained as:
\begin{equation} \setlength\abovedisplayskip{6pt} \setlength\belowdisplayskip{6pt}
\mathcal{L}_E = smooth_{L1}\left ( \begin{bmatrix}
w & h
\end{bmatrix}, 
\begin{bmatrix}
w_t & h_t
\end{bmatrix}
\right )
\end{equation}

Note that $\mathcal{L}_W$ and $\mathcal{L}_E$ merely refines the width and height of boxes, without involving the angle regression.

\begin{table*}[t]
\fontsize{8.5pt}{10pt}\selectfont
\setlength{\tabcolsep}{1.2mm}
\setlength{\aboverulesep}{0.4ex}
\setlength{\belowrulesep}{0.4ex}
\setlength{\abovecaptionskip}{1.5mm}
\centering
\begin{tabular}{l|c|ccccccccccccccc|c}
\toprule
\textbf{Methods}  & * & \textbf{PL}$^1$    & \textbf{BD}    & \textbf{BR}    & \textbf{GTF}   & \textbf{SV}    & \textbf{LV}    & \textbf{SH}    & \textbf{TC}    & \textbf{BC}    & \textbf{ST}    & \textbf{SBF}   & \textbf{RA}    & \textbf{HA}    & \textbf{SP}    & \textbf{HC}    & \textbf{AP}$_\text{50}$  \\ \hline
\rowcolor{gray!20} \multicolumn{18}{l}{$\blacktriangledown$ \textit{RBox-supervised OOD}} \\ \hline
RepPoints (2019) \cite{yang2019reppoints} & \checkmark & 86.7  & 81.1  & 41.6  & 62.0  & 76.2  & 56.3  & 75.7  & 90.7  & 80.8  & 85.3  & 63.3 & 66.6  & 59.1  & 67.6  & 33.7  & 68.45 \\ 
RetinaNet (2017) \cite{lin2017focal} & \checkmark & 88.2  & 77.0  & 45.0  & 69.4  & 71.5  & 59.0  & 74.5  & 90.8  & 84.9  & 79.3  & 57.3 & 64.7  & 62.7  & 66.5  & 39.6  & 68.69 \\
GWD (2021) \cite{yang2021rethinking} & \checkmark & 89.3  & 75.4  & 47.8  & 61.9  & 79.5  & 73.8  & 86.1  & 90.9  & 84.5  & 79.4  & 55.9 & 59.7  & 63.2  & 71.0  & 45.4  & 71.66 \\
FCOS (2019) \cite{tian2019fcos} & \checkmark & 89.1  & 76.9  & 50.1  & 63.2  & 79.8  & 79.8  & 87.1  & 90.4  & 80.8  & 84.6  & 59.7 & 66.3  & 65.8  & 71.3  & 41.7  & 72.44 \\
S$^2$A-Net (2022) \cite{han2022align} & \checkmark & 89.2  & 83.0  & 52.5  & 74.6  & 78.8  & 79.2  & 87.5  & 90.9  & 84.9  & 84.8  & 61.9 & 68.0  & 70.7  & 71.4  & 59.8  & \textbf{75.81} \\ \hline
\rowcolor{gray!20} \multicolumn{18}{l}{$\blacktriangledown$ \textit{HBox-supervised OOD}} \\ \hline
Sun et al. (2021) \cite{sun2021oriented} & $\times$ & 51.5  & 38.7  & 16.1  & 36.8  & 29.8  & 19.2  & 23.4  & 83.9  & 50.6  & 80.0  & 18.9 & 50.2  & 25.6  & 28.7  & 25.5  & 38.60 \\
BoxInst-RBox (2021) \cite{tian2021boxinst}$^2$ & $\times$ & 68.4 & 40.8 & 33.1 & 32.3 & 46.9 & 55.4 & 56.6 & 79.5 & 66.8 & 82.1 & 41.2 & 52.8 & 52.8 & 65.0 & 30.0 & 53.59 \\
H2RBox (2023) \cite{yang2023h2rbox} & \checkmark & 88.5  & 73.5  & 40.8  & 56.9  & 77.5  & 65.4  & 77.9  & 90.9  & 83.2  & 85.3  & 55.3 & 62.9  & 52.4  & 63.6  & 43.3  & 67.82 \\
EIE-Det (2024) \cite{wang2024explicit}  & \checkmark & 87.7 & 70.2 & 41.5 & 60.5 & 80.7 & 76.3 & 86.3 & 90.9 & 82.6 & 84.7 & 53.1 & 64.5 & 58.1 & 70.4 & 43.8 & 70.10 \\
H2RBox-v2 (2023) \cite{yu2023h2rboxv2} &\checkmark & 89.0 & 74.4 & 50.0 & 60.5 & 79.8 & 75.3 & 86.9 & 90.9 & 85.1 & 85.0 & 59.2 & 63.2 & 65.2 & 70.5 & 49.7 & \textbf{72.31} \\ \hline
\rowcolor{gray!20} \multicolumn{18}{l}{$\blacktriangledown$ \textit{Point-supervised OOD}} \\ \hline
\footnotesize Point2Mask-RBox (2023) \cite{li2023point2mask}$^2$  & $\times$ & 4.0 & 23.1 & 3.8 & 1.3 & 15.1 & 1.0 & 3.3 & 19.0 & 1.0 & 29.1 & 0.0 & 9.5 & 7.4 & 21.1 & 7.1 & 9.72 \\
\scriptsize P2BNet+H2RBox (2023) \cite{chen2022pointtobox,yang2023h2rbox} & $\times$ & 24.7 & 35.9 & 7.1 & 27.9 & 3.3 & 12.1 & 17.5 & 17.5 & 0.8 & 34.0 & 6.3 & 49.6 & 11.6 & 27.2 & 18.8 & 19.63 \\
\scriptsize P2BNet+H2RBox-v2 (2023) \cite{chen2022pointtobox,yu2023h2rboxv2} & $\times$ & 11.0 & 44.8 & 14.9 & 15.4 & 36.8 & 16.7 & 27.8 & 12.1 & 1.8 & 31.2 & 3.4 & \underline{50.6} & 12.6 & 36.7 & 12.5 & 21.87\\
P2RBox (2024) \cite{cao2023p2rbox}$^\dagger$ & $\times$ & \underline{87.8} & \underline{65.7} & \underline{15.0} & \textbf{60.7} & \underline{73.0} & \underline{71.7} & \underline{78.9} & 81.5 & 44.5 & \underline{81.2} & \textbf{41.2} & 39.3 & \underline{45.5} & \underline{57.5} & \underline{41.2} & \underline{59.04} \\ 
PointOBB (2024) \cite{luo2024pointobb} & $\times$ & 26.1 & \underline{65.7} & 9.1 & \underline{59.4} & 65.8 & 34.9 & 29.8 & 0.5 & 2.3 & 16.7 & 0.6 & 49.0 & 21.8 & 41.0 & 36.7  & 30.08 \\ 
\footnotesize Point2RBox (2024) \cite{yu2024point2rbox} & \checkmark & 62.9 & 64.3 & 14.4 & 35.0 & 28.2 & 38.9 & 33.3 & 25.2 & 2.2  & 44.5 & 3.4  & 48.1 & 25.9 & 45.0 & 22.6 & 34.07 \\
\footnotesize Point2RBox+SK (2024) \cite{yu2024point2rbox}$^\dagger$ & \checkmark & 53.3 & 63.9 & 3.7  & 50.9 & 40.0 & 39.2 & 45.7 & 76.7 & 10.5 & 56.1 & 5.4  & 49.5 & 24.2 & 51.2 & 33.8 & 40.27 \\
\footnotesize Point2RBox+SK (2024) \cite{yu2024point2rbox}$^\dagger$ & $\times$ & 66.4 & 59.5 & 5.2 & 52.6 & 54.1 & 53.9 & 57.3 & \textbf{90.8} & 3.2 & 57.8 & 6.1  & 47.4 & 22.9 & 55.7 & 40.5 & 44.90 \\
PointOBB-v2 (2025) \cite{ren2024pointobbv2} & $\times$ & 64.5 & 27.8 & 1.9 & 36.2 & 58.8 & 47.2 & 53.4 & \underline{90.5} & 62.2 & 45.3 & 12.1 & 41.7 & 8.1 & 43.7 & 32.0 & 41.68 \\ 
PointOBB-v3 (2025) \cite{zhang2025pointobbv3} & \checkmark & 30.9 & 39.4 & 13.5 & 22.7 & 61.2 & 7.0 & 43.1 & 62.4 & 59.8 & 47.3 & 2.7 & 45.1 & 16.8 & 55.2 & 11.4 & 41.29 \\
PointOBB-v3 (2025) \cite{zhang2025pointobbv3} & $\times$ & 52.9 & 54.4 & \textbf{21.3} & 52.7 & 65.6 & 44.9 & 67.8 & 87.2 & 26.7 & 73.4 & \underline{32.6} & \textbf{53.3} & 39.0 & 56.4 & 10.2 & 49.24 \\ 
\rowcolor{gray!20} Point2RBox-v2 (ours) & \checkmark & 78.4 & 52.7 & 8.3 & 40.9 & 71.0 & 60.5 & 74.7 & 88.7 & \underline{65.5} & 72.1 & 24.4 & 26.1 & 30.1 & 50.7 & 21.0 & 51.00 \\ 
\rowcolor{gray!20} Point2RBox-v2 (ours) & $\times$ & \textbf{88.0} & \textbf{72.6} & 8.0 & 46.2 & \textbf{79.6} & \textbf{76.3} & \textbf{86.9} & 89.1 & \textbf{79.7} & \textbf{82.9} & 26.2 & 45.3 & \textbf{45.8} & \textbf{66.3} & \textbf{46.3} & \textbf{62.61} \\ \bottomrule
\specialrule{0pt}{2pt}{0pt}
\multicolumn{18}{l}{$^*$Comparison tracks: \checkmark = End-to-end training and testing; $\times$ = Generating pseudo labels to train the FCOS detector (two-stage training).} \\
\multicolumn{18}{l}{$^\dagger$Using additional priors. P2RBox: Pre-trained SAM model; Point2RBox+SK: One-shot sketches for each class.} \\
\multicolumn{18}{l}{$^1$PL: Plane, BD: Baseball diamond, BR: Bridge, GTF: Ground track field, SV: Small vehicle, LV: Large vehicle, SH: Ship, TC: Tennis court,} \\
\multicolumn{18}{l}{$\,\;$BC: Basketball court, ST: Storage tank, SBF: Soccer-ball field, RA: Roundabout, HA: Harbor, SP: Swimming pool, HC: Helicopter.} \\
\multicolumn{18}{l}{$^2$-RBox: The minimum rectangle operation is performed on the output Mask to obtain the RBox.} \\
\bottomrule
\end{tabular}
\caption{Detection performance of each category on the DOTA-v1.0 and the mean AP$_\text{50}$ of all categories.}
\label{tab:exp_dota}
\vspace{-10pt}
\end{table*}

\subsection{Symmetry-aware Learning}
\label{sec:method-lss}

Proven effective in learning the rotation of objects \cite{yu2023h2rboxv2}, in this work, symmetry-aware learning is extended to Gaussian-based OOD. In the left part of Fig.~\ref{fig:arch}, the training image $I$ is transformed (randomly selected from rotation, flip, and scale, see Table \ref{tab:abl_pro}) to generate an augmented view as:
\begin{equation} \setlength\abovedisplayskip{6pt} \setlength\belowdisplayskip{6pt}
I_\text{aug} = \alpha I
\end{equation}
where
\begin{equation} \setlength\abovedisplayskip{6pt} \setlength\belowdisplayskip{6pt}
\alpha = \begin{bmatrix}
 \cos \mathcal{R} & -\sin \mathcal{R} \\
 \sin \mathcal{R} & \cos \mathcal{R}
\end{bmatrix}^{p_1}
\begin{bmatrix}
 1 & 0 \\
 0 & -1
\end{bmatrix}^{p_2}
\begin{bmatrix}
 s & 0 \\
 0 & s
\end{bmatrix}^{p_3}
\end{equation}
where $(p_1, p_2, p_3) = (1, 0, 0)$ when rotation is selected, $(0, 1, 0)$ when flip, $(0, 0, 1)$ when scale. $\mathcal{R}$ and $s$ are the random amount of rotation and scale, $s \in (0.5, 0.9)$.

By feeding both $I$ and $I_\text{aug}$ into the network, we obtain two sets of output Gaussian distributions and angles:
\begin{equation} \setlength\abovedisplayskip{6pt} \setlength\belowdisplayskip{6pt}
\left\{\begin{array}l
\left ( \Sigma, \theta
\right ) = f_\text{nn}\left ( I \right ) \\
\left ( \Sigma_\text{aug}, \theta_\text{aug}
\right ) = f_\text{nn}\left ( I_\text{aug} \right )
\end{array}\right.
\end{equation}

The consistency loss is calculated between the two sets so that the network learns the size and rotation variation:
\begin{equation} \setlength\abovedisplayskip{6pt} \setlength\belowdisplayskip{6pt}
\mathcal{L}_\text{ss} = L_\text{GWD}\left ( \alpha\Sigma\alpha^\top, \Sigma_\text{aug} \right ) 
\!+\! L_\text{ANG}\left ( m\theta \!+\! \mathcal{R}, \theta_\text{aug} \right )
\end{equation}
where $\mathcal{R}$ is the rotation angle and $m=1$ when rotation is selected; $(m,\mathcal{R})=(-1,0)$ when flip; $(m,\mathcal{R})=(1,0)$ when scale. $L_\text{GWD}\left (\cdot\right )$ is Gaussian Wasserstein Distance Loss \cite{yang2023detecting} and $L_\text{ANG}\left (\cdot\right )$ is defined as:
\begin{equation} \setlength\abovedisplayskip{6pt} \setlength\belowdisplayskip{6pt}
L_\text{ANG}\left ( \theta_1, \theta_2 \right ) =\min_{k\in Z} \left ( smooth_{L1}\left ( \theta_1, k\pi+\theta_2  \right ) \right )
\end{equation}
where $\min\left ( \cdot \right )$ regresses the prediction toward the closest target to circumvent the periodicity problem \cite{yu2023h2rboxv2}.

\subsection{Copy-paste Augmentation}
\label{sec:method-cp}

Inspired by \cite{ghiasi2021copypaste}, we propose to crop the detected instances of step $k$ and paste them on the training image of step $k + 1$. The maximum number of paste boxes is limited to 10 in each step. We simply use the bounding boxes of the cropped instances as the regression targets, and use Gaussian Wasserstein Distance Loss \cite{yang2023detecting} to calculate $\mathcal{L}_\text{box}$.

\subsection{Overall Loss}

The overall loss $\mathcal{L}$ for Point2RBox-v2 can be expressed as:
\begin{equation} \setlength\abovedisplayskip{6pt} \setlength\belowdisplayskip{6pt}
\mathcal{L}_\text{cls}\! +\! w_\text{box} \mathcal{L}_\text{box}\! +\! w_\text{O} \mathcal{L}_\text{O}\! +\! w_\text{W} \mathcal{L}_\text{W}\! +\! w_\text{E} \mathcal{L}_\text{E}\! +\! w_\text{ss} \mathcal{L}_\text{ss}
\end{equation}
where $\mathcal{L}_\text{cls}$ is the focal loss \cite{lin2017focal} for classification, $\mathcal{L}_\text{box}$ regresses boxes/centers toward copy-paste/ground-truth labels, $w_\text{box}$ is set to one by default, $(w_\text{O},w_\text{W},w_\text{E},w_\text{ss})$ are set to $(10, 5, 0.3, 1)$ based on our ablation studies (see Tables \ref{tab:abl_lo}-\ref{tab:abl_le}).

\section{Experiments}
\label{sec:experiment}

Experiments are carried out on NVIDIA RTX4090 GPUs using PyTorch 2.2.0 \cite{paszke2019pytorch} and the rotation detection tool kits: MMRotate 1.0.0 \cite{zhou2022mmrotate}. All the experiments follow the same hyper-parameters (learning rate, batch size, optimizer, etc.).

Average precision (AP) is adopted as the primary metric. All the models are configured upon ResNet50 \cite{he2016deep} and trained with AdamW \cite{loshchilov2018decoupled}.
\textbf{1) Learning rate.} Initialized at 5e-5, warm-up for 500 iterations, and divided by ten at each decay step. 
\textbf{2) Epochs.} 72 for HRSC; 12 for the others.
\textbf{3) Augmentation.} Random rotation/flip for HRSC; random flip for the others.
\textbf{4) Image size.} Split into 1,024 $\times$ 1,024 with an overlap of 200 for DOTA/FAIR1M/STAR; scaled to 800 $\times$ 800 for others.
\textbf{5) Multi-scale.} All experiments evaluated without multi-scale technique \cite{zhou2022mmrotate}. 
\textbf{6) Datasets.} Six remote sensing and one retail scene datasets, covering all datasets used by the main counterparts \cite{yu2024point2rbox, luo2024pointobb, cao2023p2rbox}:

\begin{table*}[!tb]
\fontsize{8.5pt}{10pt}\selectfont
\setlength{\tabcolsep}{0.65mm}
\setlength{\aboverulesep}{0.4ex}
\setlength{\belowrulesep}{0.4ex}
\setlength{\abovecaptionskip}{1.5mm}
\centering
\begin{tabular}{l|c|c|c|c|c|c|c|c|c|c}
\toprule
{\textbf{Methods}} & {*} & {\textbf{\,DOTA-v1.0\,}} & {\textbf{\,DOTA-v1.5\,}} & {\textbf{\,DOTA-v2.0\,}} & {\textbf{~~DIOR~~}} & {\textbf{~~HRSC~~}} & {\textbf{\,FAIR1M\,}} & {\textbf{~~STAR~~}} & {\textbf{\,SKU110K\,}} & {\textbf{~~RSAR~~}} \\
\hline
\rowcolor{gray!20} \multicolumn{11}{l}{$\blacktriangledown$ \textit{RBox-supervised OOD}} \\ \hline
RetinaNet (2017) \cite{lin2017focal} & \checkmark & 68.69 & 60.57        & 47.00 & 54.96 & 84.49   & 37.67   & 21.80 & 78.50 & 57.67  \\
GWD (2021) \cite{yang2021rethinking} & \checkmark & 71.66 & 63.27        & 48.87 & 57.60 & 86.67   & 39.11   & 25.30 & 79.16 & 57.80 \\
FCOS (2019) \cite{tian2019fcos} & \checkmark & 72.44 & 64.53        & 51.77    &  59.83  & 88.99  & 41.25   & \textbf{28.10} & 80.09 & \textbf{66.66} \\
S$^2$A-Net (2022) \cite{han2022align} & \checkmark & \textbf{75.81} & \textbf{66.53} & \textbf{52.39} & \textbf{61.41} & \textbf{90.10} & \textbf{42.44}   & 27.30 & \textbf{80.36} & 66.47 \\
\hline
\rowcolor{gray!20} \multicolumn{11}{l}{$\blacktriangledown$ \textit{HBox-supervised OOD}} \\ \hline
Sun et al. (2021) \cite{sun2021oriented} & $\times$ & 38.60 & - & - & - & - & - & - & - & - \\
KCR (2023) \cite{zhu2023knowledge} & \checkmark & - & - & - & - &  79.10  & -  & - & - & -  \\
H2RBox (2023) \cite{yang2023h2rbox} & \checkmark & 70.05 & 61.70        & 48.68    & 57.80 &  7.03  & 35.94  & 17.20 & 57.15 & 49.92    \\
H2RBox-v2 (2023) \cite{yu2023h2rboxv2} & \checkmark & 72.31 & 64.76 & 50.33 & 57.64 & \textbf{89.66} & \textbf{42.27} & \textbf{27.30} & \textbf{70.70} & \textbf{65.16} \\
AFWS (2024) \cite{lu2024afws} & \checkmark & \textbf{72.55} & \textbf{65.92} & \textbf{51.73} & \textbf{59.07} & - & 41.80 & - & - & - \\
\hline
\rowcolor{gray!20} \multicolumn{11}{l}{$\blacktriangledown$ \textit{Point-supervised OOD}} \\ \hline
P2RBox (2024) \cite{cao2023p2rbox}$^\dagger$ & $\times$ & \underline{59.04} & -        & - & - & -   & -  & -  & - & -  \\
PointSAM (2024) \cite{liu2024pointsam}$^\dagger$ & $\times$ & - & - & - & \textbf{46.20} & -   & -  & -  & - & - \\
PointOBB (2024) \cite{luo2024pointobb} & $\times$ & 30.08 & 10.66        & 5.53     &  37.31  & -   & 11.19 & 9.19  & - & 13.80    \\
Point2RBox+SK (2024) \cite{yu2024point2rbox}$^\dagger$ & \checkmark & 40.27 & 30.51        & 23.43    & 27.34 & 79.40   & 20.03 & 7.86  & 3.41 & 27.81    \\
PointOBB-v2 (2025) \cite{ren2024pointobbv2} & $\times$ & 41.68 & 30.59        & 20.64    &  39.56  & -   & 13.36 & 9.00  & 56.63 & 18.99   \\
PointOBB-v3 (2025) \cite{zhang2025pointobbv3} & $\checkmark$ & 41.20 & 31.25 & 22.82 & 37.60 & - & 11.42  & 11.31 & - & 15.84 \\
PointOBB-v3 (2025) \cite{zhang2025pointobbv3} & $\times$ & 49.24 & 33.79 & 23.52 & 40.18 & - & 18.35 & \underline{12.85} & - & 22.60 \\
\rowcolor{gray!20} Point2RBox-v2 (ours) & \checkmark & 51.00 & \underline{39.45} & \underline{27.11} & 34.70 & \underline{82.67} & \underline{25.72} & 7.80 & \underline{64.00} & \underline{28.60}
 \\
\rowcolor{gray!20} Point2RBox-v2 (ours) & $\times$ & \textbf{62.61} & \textbf{54.06}        & \textbf{38.79}   & \underline{44.45}  & \textbf{86.15}   & \textbf{34.71}  & \textbf{14.20} & \textbf{65.64} & \textbf{30.90}    \\
\bottomrule
\specialrule{0pt}{2pt}{0pt}
\multicolumn{11}{l}{$^*$Comparison tracks: \checkmark = End-to-end training and testing; $\times$ = Generating pseudo labels to train the FCOS detector (two-stage training).} \\
\multicolumn{11}{l}{$^\dagger$Using additional priors. P2RBox/PointSAM: Pre-trained SAM model; Point2RBox+SK: One-shot sketches for each class.} \\
\bottomrule
\end{tabular}
\caption{Accuracy (AP$_{50}$) comparisons on the DOTA-v1.0/1.5/2.0, DIOR, HRSC, FAIR1M, STAR, SKU110K, and RSAR datasets.}
\label{tab:exp_other}
\vspace{-4pt}
\end{table*}

\begin{itemize}
    \item \textbf{DOTA \cite{xia2018dota}.} DOTA-v1.0 has 2,806 aerial images annotated with 15 categories, while DOTA-v1.5/2.0 are the extended versions with more small objects and categories.
    
    \item \textbf{DIOR \cite{cheng2022anchor}.} It is an aerial image dataset re-annotated with RBoxes based on its original HBox version \cite{li2020object}, with a high variation in object size and high intra‐class diversity. 

    \item \textbf{HRSC \cite{liu2017hrsc}.} It contains ship instances on the sea and inshore. The train/val/test set includes 436/181/444 images.

    \item \textbf{FAIR1M \cite{sun2022fair1m}.} It has more than 1 million instances and more than 40,000 images for fine-grained object recognition in remote sensing imagery, annotated with 37 categories. The results are evaluated on FAIR1M-1.0.

    \item \textbf{STAR \cite{li2024star}.} It is extensive for scene graph generation, covering more than 210,000 objects with diverse spatial resolutions, classified into 48 fine-grained categories and precisely annotated with oriented bounding boxes. 

    \item \textbf{SKU110K \cite{pan2020dynamic}.} It focuses on the detection of densely packed retail scenes with 110,712 objects in 11,762 images. The density reaches 86 instances per image. 

    \item \textbf{RSAR \cite{zhang2025rsar}.} It is a remote sensing dataset based on Synthetic Aperture Radar (SAR) imagery with 6 categories.

\end{itemize}

\begin{table*}[!tb]
\fontsize{8.5pt}{10pt}\selectfont
\setlength{\tabcolsep}{2.08mm}
\setlength{\aboverulesep}{0.4ex}
\setlength{\belowrulesep}{0.4ex}
\setlength{\abovecaptionskip}{1.5mm}
\hspace{1pt}
\begin{minipage}[t]{0.315\linewidth}
\centering
\begin{tabular}{c|cc|cc}
\toprule
\multirow{2}{*}{$w_\text{O}$} & \multicolumn{2}{c|}{\textbf{DOTA}} & \multicolumn{2}{c}{\textbf{HRSC}} \\
                  & {E2E} & {FCOS} & {E2E} & {FCOS} \\ \midrule
3  & 48.76 & 61.62 & 81.85 & 84.36 \\
5  & 49.81 & 62.44 & 82.46 & 85.76 \\
\rowcolor{gray!20} 10 & \textbf{51.00} & \textbf{62.61} & \textbf{82.67} & \textbf{86.15} \\
30 & 45.88 & 57.83 & 81.56 & 85.61 \\
\bottomrule
\end{tabular}
\caption{Ablation with the weight of $\mathcal{L}_\text{O}$.}
\label{tab:abl_lo}
\end{minipage}
\quad
\begin{minipage}[t]{0.315\linewidth}
\centering
\begin{tabular}{c|cc|cc}
\toprule
\multirow{2}{*}{$w_\text{W}$} & \multicolumn{2}{c|}{\textbf{DOTA}} & \multicolumn{2}{c}{\textbf{HRSC}} \\
                  & {E2E} & {FCOS} & {E2E} & {FCOS} \\ \midrule
3  & 50.85 & 56.78 & 78.42 & 83.49 \\
\rowcolor{gray!20} 5  & \textbf{51.00} & \textbf{62.61} & \textbf{82.67} & \textbf{86.15} \\
10 & 49.15 & 60.54 & 30.37 & 35.13 \\
30 & 42.84 & 52.53 & 23.89 & 25.91 \\
\bottomrule
\end{tabular}
\caption{Ablation with the weight of $\mathcal{L}_\text{W}$.}
\label{tab:abl_lw}
\end{minipage}
\quad
\begin{minipage}[t]{0.315\linewidth}
\setlength{\tabcolsep}{2.04mm}
\centering
\begin{tabular}{c|cc|cc}
\toprule
\multirow{2}{*}{$w_\text{E}$} & \multicolumn{2}{c|}{\textbf{DOTA}} & \multicolumn{2}{c}{\textbf{HRSC}} \\
                  & {E2E} & {FCOS} & {E2E} & {FCOS} \\ \midrule
0.1 & 48.75 & 57.62 & 34.71 & 39.45 \\
\rowcolor{gray!20} 0.3 & 51.00 & 62.61 & \textbf{82.67} & \textbf{86.15} \\
0.5 & \textbf{51.36} & \textbf{62.63} & 76.85 & 85.22 \\
1.0 & 49.05 & 60.63 & 56.59 & 59.59 \\
\bottomrule
\end{tabular}
\caption{Ablation with the weight of $\mathcal{L}_\text{E}$.}
\label{tab:abl_le}
\end{minipage}
\vspace{-4pt}
\end{table*}

\begin{table*}[!tb]
\fontsize{8.5pt}{10pt}\selectfont
\setlength{\tabcolsep}{2.04mm}
\setlength{\aboverulesep}{0.4ex}
\setlength{\belowrulesep}{0.4ex}
\setlength{\abovecaptionskip}{1.5mm}
\hspace{1pt}
\begin{minipage}[t]{0.315\linewidth}
\centering
\begin{tabular}{c|cc|cc}
\toprule
\multirow{2}{*}{$w_\text{ss}$} & \multicolumn{2}{c|}{\textbf{DOTA}} & \multicolumn{2}{c}{\textbf{HRSC}} \\
                  & {E2E} & {FCOS} & {E2E} & {FCOS} \\ \midrule
0.1 & 49.28 & 59.66 & 73.66 & 78.92 \\
\rowcolor{gray!20} 1.0 & \textbf{51.00} & \textbf{62.61} & \textbf{82.67} & \textbf{86.15} \\
3.0 & 49.15 & 59.20 & 1.30  & 1.65 \\
\bottomrule
\end{tabular}
\caption{Ablation with the weight of $\mathcal{L}_\text{ss}$.}
\label{tab:abl_lss}
\end{minipage}
\quad
\begin{minipage}[t]{0.647\linewidth}
\setlength{\tabcolsep}{3.05mm}
\centering
\begin{tabular}{c|c|c||c|c|c}
\toprule
{R / F / S} & {\textbf{DOTA}} & {\textbf{HRSC}} & {R / F / S} & {\textbf{DOTA}} & {\textbf{HRSC}} \\
 \midrule
90\% / 10\% / 0\% & 60.42 & 85.46 & 80\% / 20\% / 0\%  & 59.46 & 84.73 \\
75\% / 0\% / 25\% & 60.79 & 86.22 & 60\% / 15\% / 25\% & 62.38 & 84.21 \\
\cellcolor{gray!20}68\% / 7\% / 25\% & \cellcolor{gray!20}\textbf{62.61} & \cellcolor{gray!20}\textbf{86.15} & 38\% / 37\% / 25\% & 45.87 & 8.56  \\
45\% / 5\% / 50\% & 60.55 & 85.34 & 40\% / 10\% / 50\% & 60.49 & 10.74 \\
\bottomrule
\end{tabular}
\caption{Ablation with the proportion of augmented views in self-supervision.}
\label{tab:abl_pro}
\end{minipage}
\vspace{-10pt}
\end{table*}

\subsection{Main Results on DOTA-v1.0}
\label{sec:experiment-main}

Table \ref{tab:exp_dota} compares Point2RBox-v2 with the state-of-the-art methods, which can be categorized into two tracks: 

\textbf{1) End-to-end training.} These methods apply the trained weakly-supervised detector directly to the test set. Without relying on priors, our approach demonstrates an improvement of 16.93\% (51.00\% vs. 34.07\%) compared to Point2RBox. Even when compared to Point2RBox+SK, which incorporates additional data-side priors (i.e. one-shot examples for each class), our method still outperforms it by 10.73\% (51.00\% vs. 40.27\%).

\textbf{2) Two-stage training.} These methods generate RBox labels on train/val sets, with which the FCOS detector is trained. In this two-stage mode, Point2RBox-v2 achieves an accuracy of 62.61\%, considerably surpassing PointOBB series. Remarkably, it even outperforms the SAM-powered method P2RBox by 3.57\% (62.61\% vs. 59.04\%).

\textbf{Class-wise analysis.} The FCOS detector trained with labels generated by Point2RBox-v2 achieves accuracy nearly equivalent to RBox-supervised FCOS across six high-density categories: SH (86.9\% vs. 87.1\%), SV (79.6\% vs. 79.8\%), LV (76.3\% vs. 79.8\%), PL (88.0\% vs. 89.1\%), ST (82.9\% vs. 84.6\%), and TC (89.1\% vs. 90.4\%). Interestingly, these six high-density categories account for 88\% of DOTA instances. By annotating these categories with points and generating RBoxes using Point2RBox-v2 while labeling the other sparse categories with RBoxes, we can significantly reduce annotation labor without sacrificing much accuracy, highlighting the valuable role our method can play.

\begin{figure*}[t!]
\setlength{\abovecaptionskip}{1.2mm}
\centering
\includegraphics[width=0.96\linewidth]{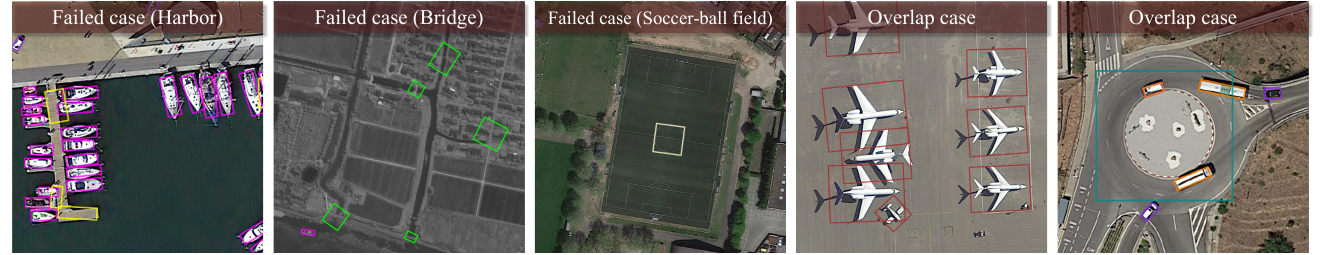}
\caption{Qualitative analysis on failed cases and overlap cases.}
\label{fig:case}
\vspace{-6pt}
\end{figure*}

\subsection{Results on More Datasets}

The results are displayed in Table \ref{tab:exp_other}.
On more challenging DOTA-v1.5/2.0, Point2RBox-v2 presents a similar trend, 23.47\%/18.15\% higher than PointOBB-v2 in the pseudo-generation track. 
On the ship detection dataset HRSC, the gap between Point2RBox-v2 and RBox-supervised FCOS is only 2.84\% (86.15\% vs. 88.99\%).
DIOR is relatively sparse, leading to less improvement with our methods---lower than PointSAM (44.45\% vs. 46.20\%) but still higher than methods that do not use SAM. 
Our method also provides competitive performance on fine-grained datasets FAIR1M and STAR. 
In addition to remote sensing scenarios, we carry out experiments on SKU110K for densely packed retail scenes. Existing point-supervised methods struggle in this case, whereas Point2RBox-v2 achieves performance on par with HBox-supervised H2RBox (65.64\% vs. 57.15\%).

\begin{table}[!tb]
\fontsize{8.5pt}{10pt}\selectfont
\setlength{\tabcolsep}{1.78mm}
\setlength{\aboverulesep}{0.4ex}
\setlength{\belowrulesep}{0.4ex}
\setlength{\abovecaptionskip}{1.5mm}
\centering
\begin{tabular}{ccccc|cc|cc}
\toprule
\multicolumn{5}{c|}{\textbf{Modules}} & \multicolumn{2}{c|}{\textbf{DOTA}} & \multicolumn{2}{c}{\textbf{HRSC}} \\
$\mathcal{L}_\text{O}$ & $\mathcal{L}_\text{W}$ & $\mathcal{L}_\text{ss}$ & $\mathcal{L}_\text{E}$ & \textit{CP} & {E2E} & {FCOS} & {E2E} & {FCOS} \\ \midrule
\checkmark & & & & & 0.00 & 0.00 & 0.00 & 0.00 \\
\checkmark & \checkmark & & & & 41.54 & 52.98 & 17.96 & 19.64 \\
\checkmark & \checkmark & \checkmark & & & 46.64 & 54.26 & 18.10 & 22.13 \\
\checkmark & \checkmark & \checkmark & \checkmark & & 49.55 & 61.88 & 78.79 & 83.79 \\
& \checkmark & \checkmark & \checkmark & \checkmark & 48.58 & 59.56 & 20.35 & 24.76 \\
\checkmark & & \checkmark & \checkmark & \checkmark & 38.94 & 48.44 & 11.64 & 14.93 \\
\checkmark & \checkmark & \checkmark & & \checkmark & 47.08 & 55.05 & 19.58 & 21.78 \\
\rowcolor{gray!20} \checkmark & \checkmark & \checkmark & \checkmark & \checkmark & \textbf{51.00} & \textbf{62.61} & \textbf{82.67} & \textbf{86.15} \\
\bottomrule
\end{tabular}
\caption{Ablation with incremental addition of modules.}
\label{tab:abl_mod}
\vspace{-4pt}
\end{table}

\begin{table}[!tb]
\fontsize{8.5pt}{10pt}\selectfont
\setlength{\tabcolsep}{2.85mm}
\setlength{\aboverulesep}{0.4ex}
\setlength{\belowrulesep}{0.4ex}
\setlength{\abovecaptionskip}{1.5mm}
\centering
\begin{tabular}{c|c|c||c|c|c}
\toprule
16 & \cellcolor{gray!20}$K\!=\!24$ & 32 & 1.2 & \cellcolor{gray!20}$\beta\!=\!1.6$ & 2.0 \\ \midrule
50.87 & \cellcolor{gray!20}\textbf{51.00} & 48.08 & 48.14 & \cellcolor{gray!20}51.00 & \textbf{51.33} \\
\bottomrule
\end{tabular}
\caption{Ablation with $K$ and $\beta$ in edge loss on DOTA (E2E).}
\label{tab:abl_edgeparam}
\vspace{-4pt}
\end{table}

\begin{table}[!tb]
\fontsize{8.5pt}{10pt}\selectfont
\setlength{\tabcolsep}{1.75mm}
\setlength{\aboverulesep}{0.4ex}
\setlength{\belowrulesep}{0.4ex}
\setlength{\abovecaptionskip}{1.5mm}
\centering
\begin{tabular}{c|cc|cc|cc}
\toprule
\multirow{2}{*}{$\sigma$} & \multicolumn{2}{c|}{Point2RBox} & \multicolumn{2}{c|}{PointOBB-v2} & \multicolumn{2}{c}{Point2RBox-v2} \\
 & {\textbf{DOTA}} & {\textbf{HRSC}} & {\textbf{DOTA}} & {\textbf{HRSC}} & {\textbf{DOTA}} & {\textbf{HRSC}} \\ \midrule
0\%  & 40.27 & 79.40 & 44.85 & - & 62.61 & 86.15 \\
10\% & 39.60 & 78.81 & 42.30 & - & 61.58 & 85.76 \\
30\% & 38.42 & 78.28 & 38.46 & - & 60.31 & 85.71 \\
\bottomrule
\end{tabular}
\caption{Ablation with the inaccuracy in point annotations.}
\label{tab:abl_noise}
\vspace{-10pt}
\end{table}

\subsection{Ablation Studies}
\label{sec:experiment-ablation}

Tables \ref{tab:abl_lo}-\ref{tab:abl_noise} display the ablation studies on DOTA-v1.0 and HRSC. ``E2E'' denotes end-to-end training; ``FCOS'' denotes two-stage training (i.e. generating pseudo labels to train FCOS). The final values adopted are highlighted in gray.

\textbf{Weight of each loss.} Tables \ref{tab:abl_lo}-\ref{tab:abl_le} determine the weights of the proposed losses. Based on these experiments, the weights $(w_\text{O},w_\text{W},w_\text{E},w_\text{ss})$ are set to $(10, 5, 0.3, 1)$.

\textbf{Proportion of augmented views.} Table \ref{tab:abl_pro} studies the proportion between rotation, flip, and scale. The results are reported with two-stage training (FCOS). Based on the results, the proportion is set to 68\%, 7\%, and 25\%.

\textbf{Incremental addition of modules.} Table \ref{tab:abl_mod} demonstrates the constraints from Gaussian and Voronoi achieve an accuracy of 52.98\% on DOTA. Adding consistency loss and edge loss further boosts it to 54.26\% and 61.88\%, respectively, whereas the improvement from copy-paste is 0.73\%. We also demonstrate the impact of omitting each core loss.

\textbf{Edge loss parameters.} We set $K=24$ and $\beta=1.6$ as they are observed to discern the correct edges during code development. Table \ref{tab:abl_edgeparam} provides a more precise ablation.

\textbf{Annotation inaccuracy.} We offset the annotated points by a noise from the uniform distribution $\left[-\sigma H, +\sigma H \right ]$, where $H$ is the height of objects. Table \ref{tab:abl_noise} shows that the AP$_{50}$ of Point2RBox-v2 decreases by less than 3\% when noise is added to point annotations, demonstrating the robustness of the proposed learning mechanisms.

\subsection{More Discussions}
\label{sec:experiment-discussions}

The qualitative analysis on the failed/overlap cases is shown in Fig. \ref{fig:case}. \textbf{1) Failed cases.} Although our method performs well overall, it struggles with certain categories that are sparse and not constrained by other objects. \textbf{2) Overlap cases.} 
Minimizing overlap as a soft constraint during training does not entirely eliminate overlap. Once trained, the model remains robust to some overlap during inference.

\section{Conclusion}
\label{sec:conclusion}

This paper introduces Point2RBox-v2, a point-supervised oriented object detector that effectively leverages the arrangement and layout of instances. We propose the integration of Gaussian overlay and Voronoi tessellation to constrain the size and rotation of instances based on their spatial relationships. Additionally, by incorporating self-supervised consistency loss, edge loss, and copy-paste augmentation, the accuracy of the model is further enhanced.

Experiments yield the following observations: 
\textbf{1)} The integration of Gaussian and Voronoi concepts effectively harnesses the spatial layout of objects, significantly enhancing point-supervised OOD. 
\textbf{2)} Point2RBox-v2 demonstrates exceptional performance in densely packed scenes (see Fig. \ref{fig:vis}), where existing methods struggle. 
\textbf{3)} Our method does not require priors (i.e. pre-trained SAM or one-shot examples) and is applicable to both end-to-end and pseudo-generation modes.
\textbf{4)} It advances the state of the art by a large amount, achieving 62.61\%, 86.15\%, and 34.71\% on the DOTA-v1.0, HRSC, and FAIR1M datasets, respectively.

\textbf{Limitations.} The gap between Point2RBox-v2 and the RBox-supervised OOD is still huge in terms of sparse categories (i.e. BR/SBF) since little constraint can be obtained from the layout between them when the objects are sparse.
{
    \small
    \bibliographystyle{ieeenat_fullname}
    \bibliography{main}

\begin{thebibliography}{73}
\providecommand{\natexlab}[1]{#1}
\providecommand{\url}[1]{\texttt{#1}}
\expandafter\ifx\csname urlstyle\endcsname\relax
  \providecommand{\doi}[1]{doi: #1}\else
  \providecommand{\doi}{doi: \begingroup \urlstyle{rm}\Url}\fi

\bibitem[Aurenhammer(1991)]{aurenhammer1991voronoi}
Franz Aurenhammer.
\newblock Voronoi diagrams—a survey of a fundamental geometric data structure.
\newblock \emph{ACM Computing Surveys}, 23\penalty0 (3):\penalty0 345–405, 1991.

\bibitem[Cao et~al.(2024)Cao, Yu, Yu, Han, Yang, Li, Jiao, and Han]{cao2023p2rbox}
Guangming Cao, Xuehui Yu, Wenwen Yu, Xumeng Han, Xue Yang, Guorong Li, Jianbin Jiao, and Zhenjun Han.
\newblock P2rbox: Point prompt oriented object detection with {SAM}.
\newblock \emph{arXiv preprint arXiv:2311.13128}, 2024.

\bibitem[Chen et~al.(2022)Chen, Yu, Han, Hassan, Wang, Li, Zhao, Shi, Han, and Ye]{chen2022pointtobox}
Pengfei Chen, Xuehui Yu, Xumeng Han, Najmul Hassan, Kai Wang, Jiachen Li, Jian Zhao, Humphrey Shi, Zhenjun Han, and Qixiang Ye.
\newblock Point-to-box network for accurate object detection via single point supervision.
\newblock In \emph{European Conference on Computer Vision}, 2022.

\bibitem[Cheng et~al.(2022)Cheng, Wang, Li, Xie, Lang, Yao, and Han]{cheng2022anchor}
Gong Cheng, Jiabao Wang, Ke Li, Xingxing Xie, Chunbo Lang, Yanqing Yao, and Junwei Han.
\newblock Anchor-free oriented proposal generator for object detection.
\newblock \emph{IEEE Transactions on Geoscience and Remote Sensing}, 2022.

\bibitem[Ding et~al.(2019)Ding, Xue, Long, Xia, and Lu]{ding2018learning}
Jian Ding, Nan Xue, Yang Long, Gui-Song Xia, and Qikai Lu.
\newblock Learning roi transformer for oriented object detection in aerial images.
\newblock In \emph{IEEE/CVF Conference on Computer Vision and Pattern Recognition}, pages 2849--2858, 2019.

\bibitem[Feng et~al.(2021)Feng, Haase-Schütz, Rosenbaum, Hertlein, Gläser, Timm, Wiesbeck, and Dietmayer]{feng2021deep}
Di Feng, Christian Haase-Schütz, Lars Rosenbaum, Heinz Hertlein, Claudius Gläser, Fabian Timm, Werner Wiesbeck, and Klaus Dietmayer.
\newblock Deep multi-modal object detection and semantic segmentation for autonomous driving: Datasets, methods, and challenges.
\newblock \emph{IEEE Transactions on Intelligent Transportation Systems}, 22\penalty0 (3):\penalty0 1341--1360, 2021.

\bibitem[Fu et~al.(2020)Fu, Chang, Zhang, Xu, Zhang, and Sun]{fu2020rotation}
Kun Fu, Zhonghan Chang, Yue Zhang, Guangluan Xu, Keshu Zhang, and Xian Sun.
\newblock Rotation-aware and multi-scale convolutional neural network for object detection in remote sensing images.
\newblock \emph{ISPRS Journal of Photogrammetry and Remote Sensing}, 161:\penalty0 294--308, 2020.

\bibitem[Gao et~al.(2022)Gao, Zhang, Wang, Guo, Zhang, He, and Zhang]{gao2022weakly}
Shuyong Gao, Wei Zhang, Yan Wang, Qianyu Guo, Chenglong Zhang, Yangji He, and Wenqiang Zhang.
\newblock Weakly-supervised salient object detection using point supervision.
\newblock In \emph{AAAI Conference on Artificial Intelligence}, pages 670--678, 2022.

\bibitem[Ghiasi et~al.(2021)Ghiasi, Cui, Srinivas, Qian, Lin, Cubuk, Le, and Zoph]{ghiasi2021copypaste}
Golnaz Ghiasi, Yin Cui, Aravind Srinivas, Rui Qian, Tsung-Yi Lin, Ekin~D. Cubuk, Quoc~V. Le, and Barret Zoph.
\newblock Simple copy-paste is a strong data augmentation method for instance segmentation.
\newblock In \emph{IEEE/CVF Conference on Computer Vision and Pattern Recognition}, pages 2918--2928, 2021.

\bibitem[Goldman et~al.(2019)Goldman, Herzig, Eisenschtat, Goldberger, and Hassner]{goldman2019precise}
Eran Goldman, Roei Herzig, Aviv Eisenschtat, Jacob Goldberger, and Tal Hassner.
\newblock Precise detection in densely packed scenes.
\newblock In \emph{IEEE/CVF Conference on Computer Vision and Pattern Recognition}, pages 5227--5236, 2019.

\bibitem[Han et~al.(2021)Han, Ding, Xue, and Xia]{han2021redet}
Jiaming Han, Jian Ding, Nan Xue, and Gui-Song Xia.
\newblock Redet: A rotation-equivariant detector for aerial object detection.
\newblock In \emph{IEEE/CVF Conference on Computer Vision and Pattern Recognition}, pages 2785--2794, 2021.

\bibitem[Han et~al.(2022)Han, Ding, Li, and Xia]{han2022align}
Jiaming Han, Jian Ding, Jie Li, and Gui-Song Xia.
\newblock Align deep features for oriented object detection.
\newblock \emph{IEEE Transactions on Geoscience and Remote Sensing}, 60:\penalty0 1--11, 2022.

\bibitem[He et~al.(2016)He, Zhang, Ren, and Sun]{he2016deep}
Kaiming He, Xiangyu Zhang, Shaoqing Ren, and Jian Sun.
\newblock Deep residual learning for image recognition.
\newblock In \emph{IEEE Conference on Computer Vision and Pattern Recognition}, pages 770--778, 2016.

\bibitem[He et~al.(2017)He, Gkioxari, Dollar, and Girshick]{he2017maskrcnn}
Kaiming He, Georgia Gkioxari, Piotr Dollar, and Ross Girshick.
\newblock Mask r-cnn.
\newblock In \emph{IEEE International Conference on Computer Vision}, 2017.

\bibitem[Hou et~al.(2023)Hou, Lu, Yang, Li, and Xue]{hou2022grep}
Liping Hou, Ke Lu, Xue Yang, Yuqiu Li, and Jian Xue.
\newblock G-rep: Gaussian representation for arbitrary-oriented object detection.
\newblock \emph{Remote Sensing}, 15\penalty0 (3):\penalty0 757, 2023.

\bibitem[Iqbal et~al.(2021)Iqbal, Munir, Mahmood, Ali, and Ali]{iqbal2021leveraging}
Javed Iqbal, Muhammad~Akhtar Munir, Arif Mahmood, Afsheen~Rafaqat Ali, and Mohsen Ali.
\newblock Leveraging orientation for weakly supervised object detection with application to firearm localization.
\newblock \emph{Neurocomputing}, 440:\penalty0 310--320, 2021.

\bibitem[Kirillov et~al.(2023)Kirillov, Mintun, Ravi, Mao, Rolland, Gustafson, Xiao, Whitehead, Berg, Lo, et~al.]{kirillov2023segment}
Alexander Kirillov, Eric Mintun, Nikhila Ravi, Hanzi Mao, Chloe Rolland, Laura Gustafson, Tete Xiao, Spencer Whitehead, Alexander~C Berg, Wan-Yen Lo, et~al.
\newblock Segment anything.
\newblock In \emph{IEEE/CVF International Conference on Computer Vision}, pages 4015--4026, 2023.

\bibitem[Li et~al.(2020)Li, Wan, Cheng, Meng, and Han]{li2020object}
Ke Li, Gang Wan, Gong Cheng, Liqiu Meng, and Junwei Han.
\newblock Object detection in optical remote sensing images: A survey and a new benchmark.
\newblock \emph{ISPRS Journal of Photogrammetry and Remote Sensing}, 159:\penalty0 296--307, 2020.

\bibitem[Li et~al.(2025)Li, Chen, Shu, Chen, He, Yu, and Yang]{li2025simple}
Qingyun Li, Yushi Chen, Xinya Shu, Dong Chen, Xin He, Yi Yu, and Xue Yang.
\newblock A simple aerial detection baseline of multimodal language models.
\newblock \emph{arXiv preprint arXiv:2501.09720}, 2025.

\bibitem[Li et~al.(2022)Li, Chen, Hu, and Zhu]{li2022oriented}
Wentong Li, Yijie Chen, Kaixuan Hu, and Jianke Zhu.
\newblock Oriented reppoints for aerial object detection.
\newblock In \emph{IEEE/CVF Conference on Computer Vision and Pattern Recognition}, pages 1829--1838, 2022.

\bibitem[Li et~al.(2023)Li, Yuan, Wang, Zhu, Li, Liu, and Zhang]{li2023point2mask}
Wentong Li, Yuqian Yuan, Song Wang, Jianke Zhu, Jianshu Li, Jian Liu, and Lei Zhang.
\newblock Point2mask: Point-supervised panoptic segmentation via optimal transport.
\newblock In \emph{IEEE International Conference on Computer Vision}, 2023.

\bibitem[Li et~al.(2024)Li, Wang, Wang, Yang, Luo, Wang, Deng, Wang, Sun, Li, Dang, Zhang, Yu, and Junchi]{li2024star}
Yansheng Li, Linlin Wang, Tingzhu Wang, Xue Yang, Junwei Luo, Qi Wang, Youming Deng, Wenbin Wang, Xian Sun, Haifeng Li, Bo Dang, Yongjun Zhang, Yi Yu, and Yan Junchi.
\newblock Star: A first-ever dataset and a large-scale benchmark for scene graph generation in large-size satellite imagery.
\newblock \emph{arXiv preprint arXiv:2406.09410}, 2024.

\bibitem[Liao et~al.(2018)Liao, Zhu, Shi, Xia, and Bai]{liao2018rotation}
Minghui Liao, Zhen Zhu, Baoguang Shi, Gui-Song Xia, and Xiang Bai.
\newblock Rotation-sensitive regression for oriented scene text detection.
\newblock In \emph{IEEE/CVF Conference on Computer Vision and Pattern Recognition}, pages 5909--5918, 2018.

\bibitem[Lin et~al.(2017)Lin, Doll{\'a}r, Girshick, He, Hariharan, and Belongie]{Lin2017Feature}
Tsung-Yi Lin, Piotr Doll{\'a}r, Ross Girshick, Kaiming He, Bharath Hariharan, and Serge Belongie.
\newblock Feature pyramid networks for object detection.
\newblock In \emph{IEEE Conference on Computer Vision and Pattern Recognition}, pages 2117--2125, 2017.

\bibitem[Lin et~al.(2020)Lin, Goyal, Girshick, He, and Dollár]{lin2017focal}
Tsung-Yi Lin, Priya Goyal, Ross Girshick, Kaiming He, and Piotr Dollár.
\newblock Focal loss for dense object detection.
\newblock \emph{IEEE Transactions on Pattern Analysis and Machine Intelligence}, 42\penalty0 (2):\penalty0 318--327, 2020.

\bibitem[Liu et~al.(2020{\natexlab{a}})Liu, Ouyang, Wang, Fieguth, Chen, Liu, and Pietik{\"a}inen]{liu2020deep}
Li Liu, Wanli Ouyang, Xiaogang Wang, Paul Fieguth, Jie Chen, Xinwang Liu, and Matti Pietik{\"a}inen.
\newblock Deep learning for generic object detection: A survey.
\newblock \emph{International Journal of Computer Vision}, 128\penalty0 (2):\penalty0 261--318, 2020{\natexlab{a}}.

\bibitem[Liu et~al.(2024)Liu, Xu, Su, Zhang, and Li]{liu2024pointsam}
Nanqing Liu, Xun Xu, Yongyi Su, Haojie Zhang, and Heng-Chao Li.
\newblock Pointsam: Pointly-supervised segment anything model for remote sensing images.
\newblock \emph{arXiv preprint arXiv:2409.13401}, 2024.

\bibitem[Liu et~al.(2018)Liu, Liang, Yan, Chen, Qiao, and Yan]{liu2018fots}
Xuebo Liu, Ding Liang, Shi Yan, Dagui Chen, Yu Qiao, and Junjie Yan.
\newblock Fots: Fast oriented text spotting with a unified network.
\newblock In \emph{IEEE/CVF Conference on Computer Vision and Pattern Recognition}, pages 5676--5685, 2018.

\bibitem[Liu et~al.(2020{\natexlab{b}})Liu, Gao, Guo, Qin, Cai, and You]{liu2020data}
Yuekai Liu, Hongli Gao, Liang Guo, Aoping Qin, Canyu Cai, and Zhichao You.
\newblock A data-flow oriented deep ensemble learning method for real-time surface defect inspection.
\newblock \emph{IEEE Transactions on Instrumentation and Measurement}, 69\penalty0 (7):\penalty0 4681--4691, 2020{\natexlab{b}}.

\bibitem[Liu et~al.(2017)Liu, Yuan, Weng, and Yang]{liu2017hrsc}
Zikun Liu, Liu Yuan, Lubin Weng, and Yiping Yang.
\newblock A high resolution optical satellite image dataset for ship recognition and some new baselines.
\newblock In \emph{International Conference on Pattern Recognition Applications and Methods}, pages 324--331, 2017.

\bibitem[Loshchilov and Hutter(2018)]{loshchilov2018decoupled}
Ilya Loshchilov and Frank Hutter.
\newblock Decoupled weight decay regularization.
\newblock In \emph{International Conference on Learning Representations}, 2018.

\bibitem[Lu et~al.(2024)Lu, Hu, Zhu, Wei, and Li]{lu2024afws}
Junyan Lu, Qinglei Hu, Ruifei Zhu, Yali Wei, and Tie Li.
\newblock Afws: Angle-free weakly-supervised rotating object detection for remote sensing images.
\newblock \emph{IEEE Transactions on Geoscience and Remote Sensing}, 2024.

\bibitem[Luo et~al.(2024)Luo, Yang, Yu, Li, Yan, and Li]{luo2024pointobb}
Junwei Luo, Xue Yang, Yi Yu, Qingyun Li, Junchi Yan, and Yansheng Li.
\newblock Pointobb: Learning oriented object detection via single point supervision.
\newblock In \emph{IEEE/CVF Conference on Computer Vision and Pattern Recognition}, 2024.

\bibitem[Murrugarra-Llerena et~al.(2024)Murrugarra-Llerena, Kirsten, Zeni, and Jung]{murrugarra2024probabilistic}
Jeffri Murrugarra-Llerena, Lucas~N Kirsten, Luis~Felipe Zeni, and Claudio~R Jung.
\newblock Probabilistic intersection-over-union for training and evaluation of oriented object detectors.
\newblock \emph{IEEE Transactions on Image Processing}, 2024.

\bibitem[Pan et~al.(2020)Pan, Ren, Sheng, Dong, Yuan, Guo, Ma, and Xu]{pan2020dynamic}
Xingjia Pan, Yuqiang Ren, Kekai Sheng, Weiming Dong, Haolei Yuan, Xiaowei Guo, Chongyang Ma, and Changsheng Xu.
\newblock Dynamic refinement network for oriented and densely packed object detection.
\newblock In \emph{IEEE/CVF Conference on Computer Vision and Pattern Recognition}, pages 11207--11216, 2020.

\bibitem[Paszke et~al.(2019)Paszke, Gross, Massa, Lerer, Bradbury, Chanan, Killeen, Lin, Gimelshein, Antiga, Desmaison, Kopf, Yang, DeVito, Raison, Tejani, Chilamkurthy, Steiner, Fang, Bai, and Chintala]{paszke2019pytorch}
Adam Paszke, Sam Gross, Francisco Massa, Adam Lerer, James Bradbury, Gregory Chanan, Trevor Killeen, Zeming Lin, Natalia Gimelshein, Luca Antiga, Alban Desmaison, Andreas Kopf, Edward Yang, Zachary DeVito, Martin Raison, Alykhan Tejani, Sasank Chilamkurthy, Benoit Steiner, Lu Fang, Junjie Bai, and Soumith Chintala.
\newblock Pytorch: An imperative style, high-performance deep learning library.
\newblock In \emph{Advances in Neural Information Processing Systems}, pages 8024--8035, 2019.

\bibitem[Qian et~al.(2021)Qian, Yang, Peng, Yan, and Guo]{qian2021rsdet}
Wen Qian, Xue Yang, Silong Peng, Junchi Yan, and Yue Guo.
\newblock Learning modulated loss for rotated object detection.
\newblock In \emph{AAAI Conference on Artificial Intelligence}, pages 2458--2466, 2021.

\bibitem[Ren et~al.(2025)Ren, Yang, Yu, Luo, and Deng]{ren2024pointobbv2}
Botao Ren, Xue Yang, Yi Yu, Junwei Luo, and Zhidong Deng.
\newblock Pointobb-v2: Towards simpler, faster, and stronger single point supervised oriented object detection.
\newblock In \emph{International Conference on Learning Representations}, 2025.

\bibitem[Soria et~al.(2023)Soria, Li, Rouhani, and Sappa]{soria2023teed}
Xavier Soria, Yachuan Li, Mohammad Rouhani, and Angel~D. Sappa.
\newblock Tiny and efficient model for the edge detection generalization.
\newblock In \emph{IEEE/CVF International Conference on Computer Vision Workshops}, pages 1364--1373, 2023.

\bibitem[Sun et~al.(2022)Sun, Wang, Yan, Xu, Wang, Diao, Chen, Li, Feng, Xu, Weinmann, Hinz, Wang, and Fu]{sun2022fair1m}
Xian Sun, Peijin Wang, Zhiyuan Yan, Feng Xu, Ruiping Wang, Wenhui Diao, Jin Chen, Jihao Li, Yingchao Feng, Tao Xu, Martin Weinmann, Stefan Hinz, Cheng Wang, and Kun Fu.
\newblock Fair1m: A benchmark dataset for fine-grained object recognition in high-resolution remote sensing imagery.
\newblock \emph{ISPRS Journal of Photogrammetry and Remote Sensing}, 184:\penalty0 116--130, 2022.

\bibitem[Sun et~al.(2021)Sun, Ran, Yang, Gao, Kurozumi, Kimata, and Ye]{sun2021oriented}
Yongqing Sun, Jie Ran, Feng Yang, Chenqiang Gao, Takayuki Kurozumi, Hideaki Kimata, and Ziqi Ye.
\newblock Oriented object detection for remote sensing images based on weakly supervised learning.
\newblock In \emph{IEEE International Conference on Multimedia \& Expo Workshops}, pages 1--6, 2021.

\bibitem[Tian et~al.(2019)Tian, Shen, Chen, and He]{tian2019fcos}
Zhi Tian, Chunhua Shen, Hao Chen, and Tong He.
\newblock Fcos: Fully convolutional one-stage object detection.
\newblock In \emph{IEEE/CVF International Conference on Computer Vision}, pages 9626--9635, 2019.

\bibitem[Tian et~al.(2021)Tian, Shen, Wang, and Chen]{tian2021boxinst}
Zhi Tian, Chunhua Shen, Xinlong Wang, and Hao Chen.
\newblock Boxinst: High-performance instance segmentation with box annotations.
\newblock In \emph{IEEE/CVF Conference on Computer Vision and Pattern Recognition}, pages 5443--5452, 2021.

\bibitem[Vincent and Soille(1991)]{vincent1991watersheds}
L. Vincent and P. Soille.
\newblock Watersheds in digital spaces: an efficient algorithm based on immersion simulations.
\newblock \emph{IEEE Transactions on Pattern Analysis and Machine Intelligence}, 13\penalty0 (6):\penalty0 583--598, 1991.

\bibitem[Wang et~al.(2024)Wang, Zhan, Lin, Yu, Ding, Zhu, and Tao]{wang2024explicit}
Linfei Wang, Yibing Zhan, Xu Lin, Baosheng Yu, Liang Ding, Jianqing Zhu, and Dapeng Tao.
\newblock Explicit and implicit box equivariance learning for weakly-supervised rotated object detection.
\newblock \emph{IEEE Transactions on Emerging Topics in Computational Intelligence}, 2024.

\bibitem[Wen et~al.(2023)Wen, Cheng, Fang, and Li]{wen2023comprehensive}
Long Wen, Yu Cheng, Yi Fang, and Xinyu Li.
\newblock A comprehensive survey of oriented object detection in remote sensing images.
\newblock \emph{Expert Systems with Applications}, page 119960, 2023.

\bibitem[Wu et~al.(2022)Wu, Lei, and Peng]{wu2022pcbnet}
Hongjin Wu, Ruoshan Lei, and Yibing Peng.
\newblock Pcbnet: A lightweight convolutional neural network for defect inspection in surface mount technology.
\newblock \emph{IEEE Transactions on Instrumentation and Measurement}, 71:\penalty0 1--14, 2022.

\bibitem[Xia et~al.(2018)Xia, Bai, Ding, Zhu, Belongie, Luo, Datcu, Pelillo, and Zhang]{xia2018dota}
Gui-Song Xia, Xiang Bai, Jian Ding, Zhen Zhu, Serge Belongie, Jiebo Luo, Mihai Datcu, Marcello Pelillo, and Liangpei Zhang.
\newblock Dota: A large-scale dataset for object detection in aerial images.
\newblock In \emph{IEEE Conference on Computer Vision and Pattern Recognition}, pages 3974--3983, 2018.

\bibitem[Xie et~al.(2021)Xie, Cheng, Wang, Yao, and Han]{xie2021oriented}
Xingxing Xie, Gong Cheng, Jiabao Wang, Xiwen Yao, and Junwei Han.
\newblock Oriented r-cnn for object detection.
\newblock In \emph{IEEE/CVF International Conference on Computer Vision}, pages 3520--3529, 2021.

\bibitem[Yang and Yan(2020)]{yang2020arbitrary}
Xue Yang and Junchi Yan.
\newblock Arbitrary-oriented object detection with circular smooth label.
\newblock In \emph{European Conference on Computer Vision}, pages 677--694, 2020.

\bibitem[Yang and Yan(2022)]{yang2022arbitrary}
Xue Yang and Junchi Yan.
\newblock On the arbitrary-oriented object detection: Classification based approaches revisited.
\newblock \emph{International Journal of Computer Vision}, 130\penalty0 (5):\penalty0 1340--1365, 2022.

\bibitem[Yang et~al.(2018)Yang, Sun, Fu, Yang, Sun, Yan, and Guo]{yang2018automatic}
Xue Yang, Hao Sun, Kun Fu, Jirui Yang, Xian Sun, Menglong Yan, and Zhi Guo.
\newblock Automatic ship detection in remote sensing images from google earth of complex scenes based on multiscale rotation dense feature pyramid networks.
\newblock \emph{Remote sensing}, 10\penalty0 (1):\penalty0 132, 2018.

\bibitem[Yang et~al.(2019{\natexlab{a}})Yang, Yang, Yan, Zhang, Zhang, Guo, Sun, and Fu]{yang2019scrdet}
Xue Yang, Jirui Yang, Junchi Yan, Yue Zhang, Tengfei Zhang, Zhi Guo, Xian Sun, and Kun Fu.
\newblock Scrdet: Towards more robust detection for small, cluttered and rotated objects.
\newblock In \emph{IEEE/CVF International Conference on Computer Vision}, pages 8231--8240, 2019{\natexlab{a}}.

\bibitem[Yang et~al.(2021{\natexlab{a}})Yang, Hou, Zhou, Wang, and Yan]{yang2021dense}
Xue Yang, Liping Hou, Yue Zhou, Wentao Wang, and Junchi Yan.
\newblock Dense label encoding for boundary discontinuity free rotation detection.
\newblock In \emph{IEEE/CVF Conference on Computer Vision and Pattern Recognition}, pages 15814--15824, 2021{\natexlab{a}}.

\bibitem[Yang et~al.(2021{\natexlab{b}})Yang, Yan, Feng, and He]{yang2021r3det}
Xue Yang, Junchi Yan, Ziming Feng, and Tao He.
\newblock R3det: Refined single-stage detector with feature refinement for rotating object.
\newblock In \emph{AAAI Conference on Artificial Intelligence}, pages 3163--3171, 2021{\natexlab{b}}.

\bibitem[Yang et~al.(2021{\natexlab{c}})Yang, Yan, Qi, Wang, Zhang, and Qi]{yang2021rethinking}
Xue Yang, Junchi Yan, Ming Qi, Wentao Wang, Xiaopeng Zhang, and Tian Qi.
\newblock Rethinking rotated object detection with gaussian wasserstein distance loss.
\newblock In \emph{38th International Conference on Machine Learning}, pages 11830--11841, 2021{\natexlab{c}}.

\bibitem[Yang et~al.(2021{\natexlab{d}})Yang, Yang, Yang, Ming, Wang, Tian, and Yan]{yang2021learning}
Xue Yang, Xiaojiang Yang, Jirui Yang, Qi Ming, Wentao Wang, Qi Tian, and Junchi Yan.
\newblock Learning high-precision bounding box for rotated object detection via kullback-leibler divergence.
\newblock In \emph{Advances in Neural Information Processing Systems}, pages 18381--18394, 2021{\natexlab{d}}.

\bibitem[Yang et~al.(2022)Yang, Yan, Liao, Yang, Tang, and He]{yang2022scrdet++}
Xue Yang, Junchi Yan, Wenlong Liao, Xiaokang Yang, Jin Tang, and Tao He.
\newblock Scrdet++: Detecting small, cluttered and rotated objects via instance-level feature denoising and rotation loss smoothing.
\newblock \emph{IEEE Transactions on Pattern Analysis and Machine Intelligence}, 45\penalty0 (2):\penalty0 2384--2399, 2022.

\bibitem[Yang et~al.(2023{\natexlab{a}})Yang, Zhang, Li, Wang, Zhou, and Yan]{yang2023h2rbox}
Xue Yang, Gefan Zhang, Wentong Li, Xuehui Wang, Yue Zhou, and Junchi Yan.
\newblock H2rbox: Horizontal box annotation is all you need for oriented object detection.
\newblock \emph{International Conference on Learning Representations}, 2023{\natexlab{a}}.

\bibitem[Yang et~al.(2023{\natexlab{b}})Yang, Zhang, Yang, Zhou, Wang, Tang, He, and Yan]{yang2023detecting}
Xue Yang, Gefan Zhang, Xiaojiang Yang, Yue Zhou, Wentao Wang, Jin Tang, Tao He, and Junchi Yan.
\newblock Detecting rotated objects as gaussian distributions and its 3-d generalization.
\newblock \emph{IEEE Transactions on Pattern Analysis and Machine Intelligence}, 45\penalty0 (4):\penalty0 4335--4354, 2023{\natexlab{b}}.

\bibitem[Yang et~al.(2023{\natexlab{c}})Yang, Zhou, Zhang, Yang, Wang, Yan, Zhang, and Tian]{yang2023kfiou}
Xue Yang, Yue Zhou, Gefan Zhang, Jirui Yang, Wentao Wang, Junchi Yan, Xiaopeng Zhang, and Qi Tian.
\newblock The kfiou loss for rotated object detection.
\newblock In \emph{International Conference on Learning Representations}, 2023{\natexlab{c}}.

\bibitem[Yang et~al.(2019{\natexlab{b}})Yang, Liu, Hu, Wang, and Lin]{yang2019reppoints}
Ze Yang, Shaohui Liu, Han Hu, Liwei Wang, and Stephen Lin.
\newblock Reppoints: Point set representation for object detection.
\newblock In \emph{IEEE/CVF International Conference on Computer Vision}, pages 9656--9665, 2019{\natexlab{b}}.

\bibitem[Yu and Da(2023)]{yu2023psc}
Yi Yu and Feipeng Da.
\newblock Phase-shifting coder: Predicting accurate orientation in oriented object detection.
\newblock In \emph{IEEE/CVF Conference on Computer Vision and Pattern Recognition}, 2023.

\bibitem[Yu and Da(2024)]{yu2024boundary}
Yi Yu and Feipeng Da.
\newblock On boundary discontinuity in angle regression based arbitrary oriented object detection.
\newblock \emph{IEEE Transactions on Pattern Analysis and Machine Intelligence}, pages 1--14, 2024.

\bibitem[Yu et~al.(2023)Yu, Yang, Li, Zhou, Da, and Yan]{yu2023h2rboxv2}
Yi Yu, Xue Yang, Qingyun Li, Yue Zhou, Feipeng Da, and Junchi Yan.
\newblock H2rbox-v2: Incorporating symmetry for boosting horizontal box supervised oriented object detection.
\newblock In \emph{Advances in Neural Information Processing Systems}, 2023.

\bibitem[Yu et~al.(2024)Yu, Yang, Li, Da, Dai, Qiao, and Yan]{yu2024point2rbox}
Yi Yu, Xue Yang, Qingyun Li, Feipeng Da, Jifeng Dai, Yu Qiao, and Junchi Yan.
\newblock Point2rbox: Combine knowledge from synthetic visual patterns for end-to-end oriented object detection with single point supervision.
\newblock In \emph{IEEE/CVF Conference on Computer Vision and Pattern Recognition}, 2024.

\bibitem[Zhang et~al.(2025{\natexlab{a}})Zhang, Luo, Yang, Yu, Li, Zhou, Jia, Lu, Chen, Li, et~al.]{zhang2025pointobbv3}
Peiyuan Zhang, Junwei Luo, Xue Yang, Yi Yu, Qingyun Li, Yue Zhou, Xiaosong Jia, Xudong Lu, Jingdong Chen, Xiang Li, et~al.
\newblock Pointobb-v3: Expanding performance boundaries of single point-supervised oriented object detection.
\newblock \emph{arXiv preprint arXiv:2501.13898}, 2025{\natexlab{a}}.

\bibitem[Zhang et~al.(2024)Zhang, Long, Xu, and Mei]{zhang2024pmho}
Shun Zhang, Jihui Long, Yaohui Xu, and Shaohui Mei.
\newblock Pmho: Point-supervised oriented object detection based on segmentation-driven proposal generation.
\newblock \emph{IEEE Transactions on Geoscience and Remote Sensing}, 2024.

\bibitem[Zhang et~al.(2025{\natexlab{b}})Zhang, Yang, Li, Yang, Cheng, and Li]{zhang2025rsar}
Xin Zhang, Xue Yang, Yuxuan Li, Jian Yang, Ming-Ming Cheng, and Xiang Li.
\newblock Rsar: Restricted state angle resolver and rotated sar benchmark.
\newblock \emph{arXiv preprint arXiv:2501.04440}, 2025{\natexlab{b}}.

\bibitem[Zhao et~al.(2019)Zhao, Zheng, Xu, and Wu]{zhao2019object}
Zhong-Qiu Zhao, Peng Zheng, Shou-Tao Xu, and Xindong Wu.
\newblock Object detection with deep learning: A review.
\newblock \emph{IEEE Transactions on Neural Networks and Learning Systems}, 30\penalty0 (11):\penalty0 3212--3232, 2019.

\bibitem[Zhou et~al.(2017)Zhou, Yao, Wen, Wang, Zhou, He, and Liang]{zhou2017east}
Xinyu Zhou, Cong Yao, He Wen, Yuzhi Wang, Shuchang Zhou, Weiran He, and Jiajun Liang.
\newblock East: An efficient and accurate scene text detector.
\newblock In \emph{IEEE Conference on Computer Vision and Pattern Recognition}, pages 2642--2651, 2017.

\bibitem[Zhou et~al.(2022)Zhou, Yang, Zhang, Wang, Liu, Hou, Jiang, Liu, Yan, Lyu, et~al.]{zhou2022mmrotate}
Yue Zhou, Xue Yang, Gefan Zhang, Jiabao Wang, Yanyi Liu, Liping Hou, Xue Jiang, Xingzhao Liu, Junchi Yan, Chengqi Lyu, et~al.
\newblock Mmrotate: A rotated object detection benchmark using pytorch.
\newblock In \emph{30th ACM International Conference on Multimedia}, pages 7331--7334, 2022.

\bibitem[Zhu et~al.(2023)Zhu, Ferenczi, Purkait, Drummond, Rezatofighi, and van~den Hengel]{zhu2023knowledge}
Tianyu Zhu, Bryce Ferenczi, Pulak Purkait, Tom Drummond, Hamid Rezatofighi, and Anton van~den Hengel.
\newblock Knowledge combination to learn rotated detection without rotated annotation.
\newblock In \emph{IEEE/CVF Conference on Computer Vision and Pattern Recognition}, 2023.

\end{thebibliography}
}


\end{document}